\documentclass[11pt,twocolumn]{article}

\usepackage[margin=1.5cm,columnsep=.8cm]{geometry}

\usepackage{amsmath,amssymb,amsfonts}
\usepackage{algorithmic}
\usepackage{graphicx}
\usepackage{textcomp}
\usepackage{xcolor}
\usepackage{tabularx}
\usepackage{booktabs}
\usepackage{enumitem}
\usepackage[numbers]{natbib}
\usepackage{array}
\usepackage{url}

\newif\ifblind
\blindfalse

\title{Causal Reconstruction of Sentiment Signals from Sparse News Data}

\ifblind
\author{Author information removed for double-blind review}
\else
\author{
    Stefania Stan \thanks{Corresponding author, \texttt{stefania.stan@ubs.com}.}\\
    \texttt{UBS Business Solutions AG, Switzerland}
    \and
    Marzio Lunghi\\
    \texttt{UBS Business Solutions AG, Switzerland}
    \and
    Vito Vargetto\\
    \texttt{UBS Business Solutions AG, Switzerland}
    \and
    Claudio Ricci \\
    \texttt{UBS Business Solutions AG, Switzerland}
    \and
    Rolands Repetto \\
    \texttt{UBS Business Solutions AG, Switzerland}
    \and
    Brayden Leo \\
    \texttt{UBS AG, Singapore}
    \and
    Shao-Hong Gan \\
    \texttt{UBS AG, Singapore}
}

\fi

\date{}

\begin{document}

\maketitle

\begin{abstract}
Sentiment signals derived from sparse news are commonly used in financial analysis and technology monitoring, yet transforming raw article-level observations into reliable temporal series remains a largely unsolved engineering problem. Rather than treating this as a classification challenge, we propose to frame it as a \emph{causal signal reconstruction} problem: given probabilistic sentiment outputs from a fixed classifier, recover a stable latent sentiment series that is robust to the structural pathologies of news data such as sparsity, redundancy, and classifier uncertainty. We present a modular three-stage pipeline that (i)~aggregates article-level scores onto a regular temporal grid with uncertainty-aware and redundancy-aware weights, (ii)~fills coverage gaps through strictly causal projection rules, and (iii)~applies causal smoothing to reduce residual noise. Because ground-truth longitudinal sentiment labels are typically unavailable, we introduce a label-free evaluation framework based on signal stability diagnostics, information-preservation lag proxies, and counterfactual tests for causality compliance and redundancy robustness.

As a secondary external check, we evaluate the consistency of reconstructed signals against stock-price data for a multi-firm dataset of AI-related news titles (November 2024 to February 2026). The key empirical finding is a three-week lead lag pattern between reconstructed sentiment and price that persists across all tested pipeline configurations and aggregation regimes, a structural regularity more informative than any single correlation coefficient. Overall, the results support the view that stable, deployable sentiment indicators require careful reconstruction, not only better classifiers.

\noindent\textbf{Keywords:}
Sentiment Analysis,
Natural Language Processing,
Financial NLP,
Signal Reconstruction,
Label-Free Evaluation
Causal Filtering,
Weak Supervision.
\end{abstract}

\section{Introduction}
 \label{sec:intro} News sentiment plays an increasingly important role in financial markets, technology adoption, and competitive intelligence analysis. Yet despite significant progress in transformer-based sentiment classifiers, a fundamental challenge remains comparatively less studied than document-level sentiment classification: how to transform noisy, sparse, and redundant article-level outputs into a reliable \emph{temporal} sentiment signal. Most prior work focuses on improving per-document classification accuracy. In contrast, comparatively little attention has been paid to the problem of reconstructing a stable latent sentiment time series from such measurements. In many practical applications---monitoring technological trends, assessing competitive positioning, or informing portfolio overlays---the goal is not the sentiment of a single article but the evolution of sentiment over time. Moreover, any such signal must be \emph{strictly causal}: it may only use information available at or before each point in time, since look-ahead is impermissible in live deployment. This paper studies that reconstruction problem directly. Starting from probabilistic sentiment scores produced by a fixed classifier, we ask how to aggregate, gap-fill, and smooth these observations into a temporally consistent, causally valid signal. The challenge is compounded by three structural pathologies that arise ubiquitously in real news data: (i)
 \emph{sparsity and irregular timing}---coverage of specific topics or firms is often intermittent; (ii) \emph{redundancy}---syndicated stories appear across multiple outlets within the same period; and (iii) \emph{classifier uncertainty}---individual title-level probability vectors often carry substantial ambiguity that should propagate into downstream aggregation.
 \paragraph{Approach.} We propose a three-stage causal reconstruction framework:
 \begin{enumerate}
 \item \textbf{Aggregation} of article-level sentiment onto a regular temporal grid, with optional uncertainty-aware weighting and embedding-based redundancy control.
 \item \textbf{Causal filling} of missing periods using forward-carry projection rules with optional staleness decay.
 \item \textbf{Causal smoothing} via exponential moving averages, Kalman filters, or a Beta-Binomial conjugate smoother.
 \end{enumerate} Because ground-truth longitudinal labels are not available, we further introduce a \emph{label-free evaluation framework} covering four goals: stability reduction, information preservation, redundancy robustness, and causality compliance. As a secondary external plausibility check, we also compare reconstructed signals against stock-price dynamics.
 \paragraph{Validation hierarchy.} The evaluation proceeds in three layers, ordered by directness of connection to the paper's core claims:
 \begin{enumerate}
 \item \emph{Internal signal diagnostics} --- stability (total variation), lag introduced by smoothing, and gap-filling behavior.
 \item \emph{Counterfactual tests} --- impulse causality test (strict causal compliance) and duplicate injection test (robustness to redundancy).
 \item \emph{External market plausibility} --- consistency of the reconstructed signal with stock prices, interpreted as a secondary plausibility filter, not as primary validation.
 \end{enumerate}
 \paragraph{Contributions.}
 \begin{itemize}
 \item We formulate sentiment estimation as a \emph{signal reconstruction problem} from sparse, uncertain, and redundant textual measurements, rather than a classification task.
 \item We propose a modular causal reconstruction pipeline that combines uncertainty-aware aggregation, embedding-based redundancy control, causal gap-filling, and causal smoothing.
 \item We introduce a label-free evaluation framework based on signal stability, lag diagnostics, and counterfactual robustness tests.
 \item We provide an empirical study on AI-related news data showing that the framework produces signals with a consistent three-week lead--lag pattern relative to market data, stable across all tested configurations.
 \end{itemize}

\section{Related Work}
\label{sec:related}

\subsection{Sentiment Analysis in Text}

Sentiment analysis has been extensively studied in natural language
processing~\cite{gentzkow2019textasdata, pang2008opinion, liu2012sentiment}, with early approaches relying on lexicon-based techniques
and traditional machine learning classifiers~\cite{loughran2011liability}. More recently,
transformer-based language models such as BERT have significantly
improved sentiment classification performance by learning contextual
representations of text \cite{devlin2019bert}. Domain-adapted language
models have also been proposed for financial text analysis, including
FinBERT \cite{araci2019finbert}, which is specifically trained on
financial documents and news. These models have enabled more accurate
sentiment detection in specialized domains where vocabulary and
semantic context differ from general language usage.

However, most prior research focuses on predicting sentiment at the
level of individual documents or sentences. In many practical
applications, such as market monitoring or competitive intelligence,
the primary objective is not the sentiment of a single article but the
evolution of sentiment over time. Our work therefore focuses on the
problem of reconstructing a temporal sentiment signal from multiple
textual observations rather than improving the underlying classifier.

\subsection{Financial News Sentiment}

A large body of literature studies the relationship between textual
sentiment and financial market behaviour~\cite{tetlock2007giving,engelberg2011role,peress2014media}. Early work by \cite{tetlock2007giving} demonstrated that media tone can predict market movements, while  \cite{bollen2011twitter} showed that social
media sentiment may contain predictive information about stock market
returns. Subsequent studies have investigated how news sentiment,
analyst reports, and other textual sources influence individual, firm, investor
behaviour and asset prices and viceversa~\cite{KEARNEY2014171, kelly2018estimating}.

Building directly on this tradition, \cite{shapiro2022measuring, thorsrud2020words} developed a rigorous time-series measure of economic sentiment from newspaper articles spanning 1980–2015, demonstrating the methodological challenges of constructing stable aggregated sentiment indices from raw article-level data. More recently, \cite{bybee2024business} showed that temporal news-attention series derived from aggregated newspaper text closely track economic activity and forecast aggregate stock returns, illustrating both the informational value and the aggregation challenges that motivate the present work.

Financial sentiment research often constructs sentiment indices by
aggregating textual signals from news articles or social media posts.
These approaches typically rely on large datasets and assume dense
coverage of textual observations. In contrast, many real-world
applications involve sparse and irregular observations, especially
when monitoring specific technologies, companies, or topics. In such
settings, naive aggregation can lead to unstable sentiment indicators.
Our work addresses this issue by proposing a structured reconstruction
pipeline that explicitly accounts for sparsity, redundancy, and
uncertainty in textual measurements.

\subsection{Signal Extraction from Noisy Observations}

The problem of estimating a latent signal from noisy measurements has
been widely studied in time-series analysis. Classical approaches
include exponential smoothing \cite{brown1959forecasting, hyndman2008forecasting}, Kalman
filtering \cite{kalman1960new}, and state-space models~\cite{durbin2012statespace,hamilton1994timeseries, casals2026noisy}. Harvey's structural time series framework \cite{harvey1990forecasting} similarly provides foundational methodology for unobserved-component models with irregular observation schedules.

In the context of textual sentiment, individual observations can be
viewed as noisy measurements of an underlying latent sentiment state.
However, unlike traditional time-series settings, textual observations
are often irregularly spaced and may contain duplicate reports of the
same underlying event. Our framework adapts classical signal
extraction ideas to this setting by combining temporal aggregation,
redundancy control, and strictly causal smoothing to produce a stable
sentiment time series.

\subsection{Weak Supervision and Label-Free Evaluation}

In many practical applications, ground-truth sentiment labels are not
available for evaluating long-horizon sentiment signals. Weak
supervision methods address this challenge by combining multiple
noisy signals to produce useful estimates without relying on manually
labeled data \cite{ratner2017snorkel}. Similar ideas have been applied
in domains where direct supervision is difficult or expensive.

Our work adopts a related perspective by introducing a label-free
evaluation framework for sentiment signal reconstruction. Instead of
comparing predictions against labeled sentiment data, we evaluate the
quality of reconstructed signals using signal-based diagnostics,
including stability, lag behaviour, and robustness to counterfactual
perturbations. This approach enables systematic evaluation of
sentiment reconstruction pipelines even when ground-truth sentiment
labels are unavailable.

\subsection{Positioning of This Work}

This paper bridges natural language processing, financial text analysis, and time-series signal extraction. The precise gap it addresses is more specific than ``few papers reconstruct stable sentiment signals'': much of the literature focuses on extraction and association rather than causal reconstruction under sparse and redundant observations, treats redundancy and classifier uncertainty as separate preprocessing concerns if at all, and rarely enforces or evaluates strict causality in the resulting series.
Prior financial text studies largely focus on sentiment extraction and market association; they devote less attention to causal reconstruction under sparse, redundant, and irregular observations~\cite{KEARNEY2014171}.
The framework proposed here addresses all three simultaneously under a unified reconstruction lens, and introduces a label-free evaluation methodology that does not require ground-truth longitudinal sentiment labels.

\section{Sentiment Signal Reconstruction Framework}
\label{sec:sentiment_modeling}

News titles arrive irregularly, are sometimes duplicated across outlets, and carry heterogeneous classifier confidence. The objective of this section is to turn these raw observations into a stable, regular-grid latent sentiment series $S_S^{(p)}$ that is robust to all three pathologies and respects strict causality at every step.

The reconstruction proceeds through three sequential stages:
\begin{equation}
\label{eq::framework}
S_A^{(p)} \;\leftarrow\; S^{(p)},
\qquad
S_F^{(p)} \;\leftarrow\; S_A^{(p)},
\qquad
S_S^{(p)} \;\leftarrow\; S_F^{(p)},
\end{equation}
where $B \leftarrow A$ denotes that series $B$ is derived from series $A$ by the operator defined in the corresponding subsection. To orient the reader through the component choices that follow, we distinguish three layers:

\begin{itemize}
  \item \textbf{Core pipeline} (always present): fixed-grid aggregation
        \ref{sec:aggregation} $\to$ causal filling
        \ref{sec:filling} $\to$ causal smoothing
        \ref{sec:smoothing}.
  \item \textbf{Configurable enhancements} (optional): uncertainty-aware
        weighting $W_{\mathrm{unc}}$, embedding-based redundancy control
        $W_{\mathrm{dup}}$, intra-bin recency weighting $W_{\mathrm{time}}$,
        and the global vs.\ local aggregation choice. All are defined in
        \ref{sec:aggregation}.
  \item \textbf{Smoother variants} (compared empirically in
        \ref{sec:design_study}): EWMA, Kalman family, Beta-Binomial, and the Weighted Kalman variations.
\end{itemize}

\subsection{Notation and problem setup}
\label{sec:article_level}
Banks are indexed by $p\in\mathcal{P}$. For each bank $p$, we observe a set of news titles indexed by $i$, each with timestamp $t_{p,i}$ and a fixed sentiment classifier output
\begin{equation}
\boldsymbol{\pi}_{p,i}
=
\bigl(\pi_{p,i,\mathrm{pos}},\,\pi_{p,i,\mathrm{neg}},\,\pi_{p,i,\mathrm{neu}}\bigr),
\;
\sum_{c\in\{\mathrm{pos},\mathrm{neg},\mathrm{neu}\}} \pi_{p,i,c}=1.
\end{equation}
We map $\boldsymbol{\pi}_{p,i}$ to a scalar sentiment score
\begin{equation}
\label{eq:article_score}
s_{p,i}
=
\pi_{p,i,\mathrm{pos}}-\pi_{p,i,\mathrm{neg}}
\in[-1,1],
\end{equation}
where positive values indicate positive sentiment and negative values indicate negative sentiment.

\paragraph{Irregular sparse series.}
Articles are collected via vocabulary-based scraping, where each article belongs to one or more thematic \emph{categories} (vocabularies) $\mathcal{V}$. For each bank $p$, the raw (irregular, sparse) observation set is
\begin{equation}
\label{eq:raw_sparse_series}
S^{(p)}=\{(t_{p,i},\, s_{p,i},\, v_{p,i})\}_i,
\end{equation}
where $v_{p,i}\in\mathcal{V}$ denotes the category of article $i$. In practice, $S^{(p)}$ is sparse (coverage gaps), redundant (near-duplicate coverage), and uncertain (ambiguous classifier posteriors). We therefore apply a sequential reconstruction pipeline on a fixed aggregation grid presented in Eq.~\eqref{eq::framework}.
The final output can be any stage $S_{\mathrm{final}}^{(p)}\in\{S_A^{(p)},\,S_F^{(p)},\,S_S^{(p)}\}$; when multiple stages are used, the order above is always respected.

\subsection{Stage 1 — Aggregation: $S_A^{(p)} \leftarrow S^{(p)}$}
\label{sec:aggregation}

\subsubsection{Global vs.\ local aggregation}
\label{sec:global_local}

A key design choice is whether to aggregate articles across all categories jointly or separately per category before combining. We distinguish two strategies.

\paragraph{Global aggregation.}
All articles belonging to bank $p$, regardless of their category $v_{p,i}$, are pooled together within each time bin and reduced to a single grid series $S_A^{(p)}$. This is the most compact approach and is appropriate when the signal of interest is the overall sentiment across all thematic dimensions simultaneously.

\paragraph{Local aggregation.}
Articles are first separated by category. For each category $v\in\mathcal{V}$ and bank $p$, a category-level grid series $S_A^{(p,v)}$ is constructed independently using the aggregation mechanics described below. The per-bank series is then obtained by aggregating across categories:
\begin{equation}
\label{eq:local_aggregation}
S_A^{(p)}(k)
=
g_{\mathcal{V}}\!\bigl(\{S_A^{(p,v)}(k) : v\in\mathcal{V},\, S_A^{(p,v)}(k)\neq\mathrm{NA}\}\bigr),
\end{equation}
where $g_{\mathcal{V}}$ is a cross-category reducing function (e.g., a weighted or unweighted mean over categories with non-missing values). Local aggregation treats each vocabulary as an independent signal source, offering finer control over how different thematic dimensions contribute to the final estimate and reducing the risk that a high-volume category overwhelms lower-coverage but equally informative ones.

Both strategies share the same article-level aggregation mechanics described below; they differ only in whether the category dimension is resolved before or after bin-level reduction.

\subsubsection{Bin-level aggregation}
\label{sec:bin_aggregation}
To stabilize the signal under scarcity, we aggregate article-level scores onto a regular grid (e.g., weekly or monthly). Let $\{D_1,\dots,D_T\}$ be a partition of the time horizon into consecutive aggregation periods. For bank $p$ and bin $D_k$, define the set of article indices falling in that bin:
\begin{equation}
\label{eq:bin_index_set}
I_{p,k}=\{\, i \mid t_{p,i}\in D_k \,\}.
\end{equation}
We define the aggregated grid series $S_A^{(p)}(k)$ by
\begin{equation}
\label{eq:aggregation_def}
S_A^{(p)}(k)=
\begin{cases}
f\!\left(\{(s_{p,i},w_{p,i}) : i\in I_{p,k}\}\right), & |I_{p,k}|>0,\\
\mathrm{NA}, & |I_{p,k}|=0,
\end{cases}
\end{equation}
where $\mathrm{NA}$ denotes a missing bin (no coverage) and $w_{p,i}\ge 0$ is a (possibly composite) weight.

A practical default for $f$ is the weighted mean
\begin{equation}
\label{eq:weighted_mean}
f\!\left(\{(s_{p,i},w_{p,i})\}_{i\in I_{p,k}}\right)
=
\frac{\sum_{i\in I_{p,k}} w_{p,i}\, s_{p,i}}{\sum_{i\in I_{p,k}} w_{p,i}},
\end{equation}
with the unweighted mean recovered by $w_{p,i}\equiv 1$. Other reducing functions (e.g., trimmed means or M-estimators) can be substituted without changing the surrounding framework; we do not evaluate them here.

\subsubsection{Uncertainty scaling from $\boldsymbol{\pi}_{p,i}$}
\label{sec:uncertainty_scaling}
We downweight ambiguous titles using uncertainty derived from $\boldsymbol{\pi}_{p,i}$. We consider three uncertainty functionals, each mapping an arbitrary probability vector $\boldsymbol{\pi}$ to a scalar in $[0,1]$ (or a bounded interval), then convert uncertainty to a confidence weight. In all cases, the functional is defined generically on any valid $\boldsymbol{\pi}$ and instantiated per article as $W_{\bullet}(p,i)$.

\paragraph{Normalized entropy.}
Let $K=3$ be the number of classes. Define the normalized entropy
\begin{equation}
\label{eq:normalized_entropy}
U_{\mathrm{ent}}(\boldsymbol{\pi})
= - \frac{1}{\log K} \sum_{c=1}^{K} \pi_c \log \pi_c
\in[0,1],
\end{equation}
with higher values indicating higher uncertainty. The corresponding confidence weight is
\begin{equation}
\label{eq:conf_ent}
W_{\mathrm{ent}}(p,i)=1-U_{\mathrm{ent}}(\boldsymbol{\pi}_{p,i}) \in [0,1].
\end{equation}

\paragraph{Top-two uncertainty (margin confidence).}
Let $\pi_{(1)}$ and $\pi_{(2)}$ be the largest and second-largest entries of $\boldsymbol{\pi}_{p,i}$.
Define
\begin{equation}
\label{eq:conf_top}
W_{\mathrm{top}}(p,i)
=
\pi_{(1)}-\pi_{(2)} \in [0,1].
\end{equation}
Larger separation between the top two probabilities indicates higher classifier confidence.

\paragraph{Polarity-conflict uncertainty (three-class-specific).}
In the three-class setting $\boldsymbol{\pi}=(\pi_{\mathrm{pos}},\pi_{\mathrm{neg}},\pi_{\mathrm{neu}})$, we consider a polarity-conflict functional:
\begin{equation}
\label{eq:polarity_conflict}
U_{\mathrm{pol}}(\boldsymbol{\pi})
=
(\pi_{\mathrm{pos}}+\pi_{\mathrm{neg}})
-
(\pi_{\mathrm{pos}}-\pi_{\mathrm{neg}})^2.
\end{equation}
The term $(\pi_{\mathrm{pos}}+\pi_{\mathrm{neg}})$ measures probability mass assigned to polar sentiment, while $(\pi_{\mathrm{pos}}-\pi_{\mathrm{neg}})^2$ penalizes dominance of a single polarity. Hence $U_{\mathrm{pol}}$ is large when the classifier assigns substantial mass to non-neutral sentiment while remaining conflicted between positive and negative. Under the simplex constraint one can verify that $U_{\mathrm{pol}}\in[0,1]$, with $U_{\mathrm{pol}}=0$ when all mass is neutral or one polar class dominates entirely, and $U_{\mathrm{pol}}=1$ at maximal polar conflict ($\pi_{\mathrm{pos}}=\pi_{\mathrm{neg}}=0.5,\,\pi_{\mathrm{neu}}=0$).

The corresponding confidence weight is
\begin{equation}
\label{eq:conf_pol}
W_{\mathrm{pol}}(p,i)=1-U_{\mathrm{pol}}(\boldsymbol{\pi}_{p,i}).
\end{equation}
This downweights articles with high polar conflict and tends to upweight confidently neutral titles. At maximal polar conflict the weight is $W_{\mathrm{pol}}=0$, effectively discarding such articles from aggregation.

\paragraph{Summary.}
Depending on the configuration, we set
\begin{equation}
\label{eq:unc_choice}
W_{\mathrm{unc}}(p,i)\in\{W_{\mathrm{ent}}(p,i),\,W_{\mathrm{top}}(p,i),\,W_{\mathrm{pol}}(p,i)\},
\end{equation}
and omit uncertainty scaling by taking $W_{\mathrm{unc}}(p,i)\equiv 1$.

\begin{figure*}[!ht]
    \centering
    \includegraphics[width=0.95\linewidth]{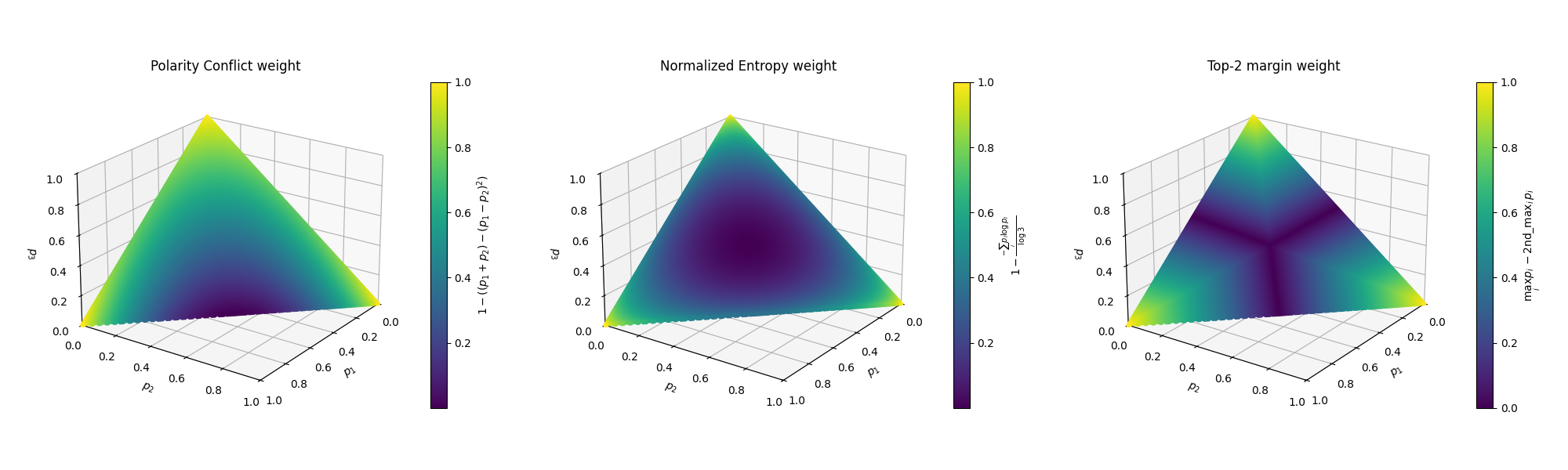}
    \caption{Visual comparison of confidence weights derived from uncertainty measures on the probability simplex.}
    \label{fig:weights}
\end{figure*}

\subsubsection{Redundancy control via embedding-based grouping}
\label{sec:near_duplicates}
News coverage often contains near-duplicate items (syndication, reposts, lightly edited rewrites). If left untreated, redundancy can over-count a single underlying story within a bin $D_k$. We therefore incorporate an optional redundancy control step \emph{within each bank $p$ and aggregation bin $D_k$}~\cite{reimers2019sbert}. Near-duplicate detection using embedding-based similarity and graph partitioning has been studied extensively in information retrieval \cite{broder1997resemblance, rodier2020online}, where connected-component grouping under cosine similarity thresholds is a standard approach to clustering syndicated documents.

For each title, compute an embedding $\mathbf{e}_{p,i}\in\mathbb{R}^d$ and measure proximity by cosine similarity
\[
\mathrm{sim}(\mathbf{e}_{p,i},\mathbf{e}_{p,j})
=
\frac{\mathbf{e}_{p,i}^\top \mathbf{e}_{p,j}}{\|\mathbf{e}_{p,i}\|_2\,\|\mathbf{e}_{p,j}\|_2}.
\]
Within each $(p,k)$, a chosen grouping method (e.g., similarity graph connected components, hierarchical clustering, or density-based clustering) maps $\{\mathbf{e}_{p,i}: i\in I_{p,k}\}$ to a partition into groups $\mathcal{C}_{p,k}=\{C_1,\dots,C_r\}$, with $C_\ell\subseteq I_{p,k}$ and singletons allowed. Let $n_i = |C(p,i)|$ denote the size of the group containing article $i$.

\paragraph{General formulation.}
The redundancy weight is defined through a scaling function $\phi:\mathbb{N}_{>0}\to\mathbb{R}_{>0}$:
\begin{equation}
\label{eq:dup_weight_general}
W_{\mathrm{dup}}(p,i) = \phi(n_i).
\end{equation}
The choice of $\phi$ encodes the analyst's belief about what redundancy signals. Two complementary interpretations lead to opposite weighting strategies.

\paragraph{Downweighting duplicates (deduplication).}
If near-duplicates are viewed as a noise artifact that should not over-represent a single story, one should reduce their collective influence. A natural family is
\begin{equation}
\label{eq:dup_weight_down}
\phi_{\mathrm{ded}}(n) = n^{-\alpha}, \qquad \alpha\in[0,1],
\end{equation}
which interpolates between no correction ($\alpha=0$) and full group normalization ($\alpha=1$, $\phi(n)=1/n$), ensuring the group contributes a single effective observation.

\paragraph{Upweighting duplicates (corroboration).}
Alternatively, if widespread multi-source coverage of the same story is interpreted as a signal of genuine salience, duplicates should be reinforced. A suitable choice is
\begin{equation}
\label{eq:dup_weight_up}
\phi_{\mathrm{cor}}(n) =
\begin{cases}
1, & n = 1,\\
\log(n), & n > 1,
\end{cases}
\end{equation}
so that singletons remain unaffected while larger groups receive a logarithmically increasing boost, preventing very large clusters from dominating the aggregation.

\paragraph{When to prefer corroboration over deduplication.}
Deduplication ($\phi_{\mathrm{ded}}$) is the conservative default: it
suppresses the potential over-counting of a single syndicated story.
Corroboration ($\phi_{\mathrm{cor}}$) is appropriate when multi-source
convergence on an event is itself informative about salience---for example,
in settings where editorial selection is noisy and repetition signals
genuine newsworthiness. In the empirical results (Section~\ref{sec:results}),
we find that global aggregation favours corroboration (cross-outlet
convergence is informative), whereas local aggregation favours deduplication
(thematic separation already limits within-category redundancy). This
distinction should be treated as a dataset-specific design choice rather
than a universal recommendation.

\subsubsection{Intra-bin recency weighting}
\label{sec:time_weight}
Articles within the same bin are not simultaneous; one may prefer more recent items within $D_k$. Let $t_k^{\mathrm{end}}$ denote the end time of bin $D_k$. A causal recency weight is
\begin{equation}
\label{eq:time_weight}
W_{\mathrm{time}}(p,i)=\exp\!\bigl(-\gamma\,(t_k^{\mathrm{end}}-t_{p,i})\bigr),
\qquad \gamma\ge 0,\ \ i\in I_{p,k}.
\end{equation}

\paragraph{Composed aggregation weights}
When enabled, weights compose multiplicatively:
\begin{equation}
\label{eq:composed_weight}
w_{p,i}
=
W_{\mathrm{unc}}(p,i)\,W_{\mathrm{dup}}(p,i)\,W_{\mathrm{time}}(p,i),
\end{equation}
where each factor can be omitted (treated as $1$) depending on the configuration. Plugging $w_{p,i}$ into Eq.~\eqref{eq:weighted_mean} yields a unified family of aggregation strategies.

\subsection{Stage 2 — Causal Filling: $S_F^{(p)} \leftarrow S_A^{(p)}$}
\label{sec:filling}

After aggregation, missing bins may arise when no articles fall within a given period. We optionally fill these gaps causally to produce a complete grid series $S_F^{(p)}$. The underlying assumption is that a missing observation does not imply the absence of a sentiment state; rather, the most recently observed sentiment persists but may grow stale over time.

Let $k^\star(k)=\max\{u\le k : S_A^{(p)}(u)\neq \mathrm{NA}\}$ be the index of the last observed bin up to $k$, and let $\Delta_k = k - k^\star(k)$ be the number of steps elapsed since that observation. Given a non-increasing decay function $g:\mathbb{N}_0\to[0,1]$ with $g(0)=1$, the filled series is
\begin{equation}
\label{eq:decayed_forward_fill}
S_F^{(p)}(k)=
\begin{cases}
S_A^{(p)}(k), & \text{if } S_A^{(p)}(k)\neq \mathrm{NA},\\
g(\Delta_k)\cdot S_A^{(p)}(k^\star(k)), & \text{otherwise.}
\end{cases}
\end{equation}
The filled value is the most recent observed sentiment scaled by a staleness factor. Two practical choices for $g$ are:

\begin{itemize}
    \item \textbf{Constant (simple forward fill).} $g(\Delta)=1$ for all $\Delta\ge 0$. The last observed value is carried forward unchanged. This assumes sentiment is persistent and does not penalize staleness.
    \item \textbf{Linear decay (finite horizon).}
    \begin{equation}
    \label{eq:decay_linear}
    g(\Delta)=\max\!\Bigl(0,\,1-\frac{\Delta}{H}\Bigr),
    \end{equation}
    where $H>0$ is the maximum carry-forward horizon in number of periods. The filled value decays linearly to zero after $H$ steps without new data, reflecting the belief that very stale estimates should not be propagated indefinitely.
\end{itemize}

Other monotonically non-increasing functions $g$ satisfying $g(0)=1$ can be substituted in Eq.~\eqref{eq:decayed_forward_fill} without modifying the surrounding framework.

\subsection{Stage 3 — Causal Smoothing: $S_S^{(p)} \leftarrow S_F^{(p)}$}
\label{sec:smoothing}

After filling, we optionally apply a strictly causal smoother to reduce residual high-frequency variability and recover a latent sentiment estimate $S_S^{(p)}$. All methods use only present and past values:
\[
S_S^{(p)}(k)=\mathcal{F}\bigl(S_F^{(p)}(1\!:\!k)\bigr),
\]
ensuring no future data leakage. In addition to the sentiment series, we denote by $t_k = |I_{p,k}|$ the article count in bin $k$, which serves as a proxy for observation reliability: bins with more articles yield more trustworthy sentiment estimates.

We organize methods into two groups: those that treat all observations equally, and those that incorporate $t_k$ to account for heteroscedastic observation uncertainty.

\subsubsection{Methods with fixed observation variance}

\paragraph{Exponentially weighted moving average (EWMA).}
\begin{equation}
\label{eq:ewma}
S_S^{(p)}(k)=\alpha\,S_F^{(p)}(k)+(1-\alpha)\,S_S^{(p)}(k-1),
\qquad \alpha\in(0,1],
\end{equation}
with initialization $S_S^{(p)}(1)=S_F^{(p)}(1)$~\cite{roberts2000control}. Higher $\alpha$ yields more responsive estimates; lower $\alpha$ produces smoother series with effective memory window approximately $1/\alpha$ observations. This method requires no distributional assumptions and is $O(1)$ per step.

\paragraph{Kalman filter (standard).}
A 1D linear Gaussian state-space model on the grid:
\begin{align}
\label{eq:kalman_state_std}
x_k &= x_{k-1} + \eta_k,
\qquad \eta_k\sim \mathcal{N}(0,\,q),\\
\label{eq:kalman_obs_std}
S_F^{(p)}(k) &= x_k + \varepsilon_k,
\qquad \varepsilon_k\sim \mathcal{N}(0,\,r),
\end{align}
where $x_k$ is the latent sentiment state, $q>0$ controls the assumed rate of signal drift, and $r>0$ is a fixed observation noise variance. The filtered estimate $S_S^{(p)}(k)=\mathbb{E}[x_k\mid S_F^{(p)}(1\!:\!k)]$ follows from standard Kalman predict--update recursions, initialized at $S_S^{(p)}(1)=S_F^{(p)}(1)$ with diffuse prior variance $P_1$. This filter is optimal under Gaussian noise and treats all observations equally regardless of bin coverage.

\paragraph{Kalman filter in arctanh space.}
Since sentiment scores lie in $[-1,1]$, a standard Gaussian state-space model does not enforce boundary constraints and may produce unbounded state estimates in extreme cases. We address this by maintaining the latent state in arctanh-transformed space:
\begin{align}
\label{eq:kalman_state_atanh}
\tilde{x}_k &= \tilde{x}_{k-1} + \eta_k,
\qquad \eta_k \sim \mathcal{N}(0,\,q),\\
\label{eq:kalman_obs_atanh}
\operatorname{arctanh}\!\bigl(S_F^{(p)}(k)\bigr) &= \tilde{x}_k + \varepsilon_k,
\qquad \varepsilon_k\sim \mathcal{N}(0,\,\tilde{r}),
\end{align}
with $\tilde{r}$ expressed in arctanh space. The smoothed output is recovered as $S_S^{(p)}(k)=\tanh(\tilde{x}_k)$, which enforces the $[-1,1]$ constraint structurally. This variant is preferred when scores concentrate near $\pm 1$.
The delta-method approximation for the observation variance in arctanh space follows standard results in asymptotic statistics \cite{vandervaart1998asymptotic}.

\subsubsection{Methods with heteroscedastic observation variance}

When article counts $t_k$ vary substantially across bins, observations from high-coverage bins should receive greater weight. The following methods incorporate $t_k$ to modulate the Kalman gain adaptively.

\paragraph{Kalman filter with count-based observation variance.}
Treating $S_F^{(p)}(k)$ as an empirical mean of $t_k$ independent scaled observations, the standard variance-of-a-proportion argument (with $s = 2p - 1$ mapping a proportion to $[-1,1]$) gives
\begin{equation}
\label{eq:hetero_var}
R_k = \frac{1 - S_F^{(p)}(k)^2}{t_k}.
\end{equation}
Substituting $R_k$ for the fixed $r$ in Eq.~\eqref{eq:kalman_obs_std} yields a heteroscedastic Kalman filter: bins with high article counts (small $R_k$) produce larger Kalman gain and stronger state updates, while sparse bins contribute proportionally less.

\paragraph{Kalman filter in arctanh space with delta-method variance.}
Combining the boundary enforcement of the arctanh transformation with heteroscedastic observation noise, the delta method yields the observation variance in arctanh space as
\begin{equation}
\label{eq:delta_var}
\tilde{R}_k = \frac{1}{t_k\,\bigl(1 - S_F^{(p)}(k)^2\bigr)}.
\end{equation}
Substituting $\tilde{R}_k$ for $\tilde{r}$ in Eq.~\eqref{eq:kalman_obs_atanh} gives a filter that simultaneously enforces the $[-1,1]$ boundary and correctly propagates heteroscedastic uncertainty. This variant is the preferred choice when scores are boundary-heavy and article counts vary widely across bins.

\paragraph{Beta-Binomial conjugate smoother.}
We re-centre $S_F^{(p)}(k)\in[-1,1]$ to a proportion $\tilde{p}_k=(S_F^{(p)}(k)+1)/2\in[0,1]$ and maintain a Beta posterior on the latent proportion with exponential forgetting:
\begin{align}
\label{eq:beta_binom_update}
\alpha_k &= \delta\,\alpha_{k-1} + \tilde{p}_k\, t_k,\\
\beta_k  &= \delta\,\beta_{k-1} + (1-\tilde{p}_k)\, t_k,
\end{align}
where $\delta\in(0,1]$ is a decay factor introducing forgetting. The smoothed sentiment estimate is mapped back as
\begin{equation}
\label{eq:beta_binom_output}
S_S^{(p)}(k) = 2\,\frac{\alpha_k}{\alpha_k+\beta_k} - 1.
\end{equation}
The filter is initialized with $\alpha_1 = \tilde{p}_1\,t_1$ and $\beta_1 = (1-\tilde{p}_1)\,t_1$. Setting $\delta=1$ recovers a cumulative posterior with no forgetting; $\delta<1$ introduces an effective memory window of approximately $1/(1-\delta)$ observations, weighted by their article counts. This approach is fully conjugate, exact, and $O(1)$ per step, making it well-suited for stationary or slowly drifting signals.

\subsubsection{Summary}

Table~\ref{tab:smoothers} summarizes the smoothing methods, their reliance on article counts, and their key properties.

\begin{table*}[!t]
\centering
\caption{Summary of causal smoothing methods. $t_k$ denotes article count in bin $k$.}
\label{tab:smoothers}
\begin{tabularx}{\textwidth}{>{\raggedright\arraybackslash}Xcc>{\raggedright\arraybackslash}X}
\hline
Method & Uses $t_k$ & Enforces $[-1,1]$ & Notes \\
\hline
EWMA & No & No & No distributional assumption; $O(1)$ \\
Kalman (standard) & No & No & Optimal under Gaussian noise \\
Kalman (arctanh) & No & Yes & Preferred near boundaries \\
Kalman (count-based $R_k$) & Yes & No & Heteroscedastic; high coverage $\to$ higher gain \\
Kalman (arctanh + delta-method $\tilde{R}_k$) & Yes & Yes & Best for boundary-heavy, variable coverage \\
Beta-Binomial & Yes & Yes (by construction) & Conjugate; suited for slow-moving signals \\
\hline
\end{tabularx}
\end{table*}

\section{Data}
\label{sec:data}

The reconstruction problem studied in this paper is motivated by three structural features that characterize real-world sparse news data: (i) irregular and intermittent article arrival, (ii) cross-outlet redundancy through syndication and paraphrase, and (iii) heterogeneous classifier confidence across titles. This section describes the dataset and shows that all three features are present.

\subsection{Collection procedure}

The dataset used in this research consists of news article titles collected through web scraping, focusing on topics related to artificial intelligence (AI). To identify relevant news, we compiled a list of AI-related terms, illustrated in table \ref{tab::terms_and_quarters}, each accompanied by a set of aliases or alternative names. These term–alias pairs were used to construct targeted search queries.

\begin{table*}[!ht]
\caption{AI Terms, Their Aliases, and Article Counts per Quarter in 2025}
\centering
\begin{tabularx}{\textwidth}{>{\raggedright\arraybackslash}X>{\raggedright\arraybackslash}Xrrrr}
\toprule
\textbf{Term} & \textbf{Aliases} & \textbf{Q1} & \textbf{Q2} & \textbf{Q3} & \textbf{Q4} \\
\midrule
AI Safety and Ethics & AI Governance, AI Alignment & 1 & 1 & 2 & 10 \\
Agentic AI & Autonomous Agent, AI Agent & 38 & 70 & 106 & 117 \\
AI Patent & AI Technology Patent, Artificial Intelligence Patent & -- & -- & -- & -- \\
Cognitive AI & Cognitive Computing & 0 & 1 & 1 & 0 \\
Conversational AI & Chatbot & 47 & 46 & 79 & 120 \\
Ethical AI & Responsible AI, Trustworthy AI, Regulate AI, AI Bubble & 34 & 58 & 95 & 200 \\
Explainable AI & Interpretable AI & 0 & 4 & 3 & 2 \\
Federated AI & Federated Learning, Collaborative AI & 0 & 0 & 2 & 0 \\
Generative AI & Gen AI, Generative Models & 169 & 208 & 269 & 285 \\
Multimodal AI & Cross-Modal AI, Multisensory AI & 0 & 1 & 1 & 2 \\
Narrow AI & Weak AI, Domain-Specific AI & 3 & 0 & 1 & 2 \\
Natural Language Processing & NLP & 8 & 9 & 21 & 12 \\
Neural Network & Artificial Neural Network & 0 & 0 & 2 & 2 \\
Neuro-Symbolic AI & Hybrid AI, Symbolic-Neural Integration & 0 & 0 & 0 & 1 \\
\bottomrule
\end{tabularx}
\label{tab::terms_and_quarters}
\end{table*}

\subsection{Sample}

Each query was further refined by specifying one of the 43 selected companies and a fixed time interval spanning from November 2024 to February 2026.
For each company, we collect news article titles published within the selected timeframe. Importantly, only the titles were retrieved, not the full article content. This decision was made for several reasons. First, article titles are generally more accessible than full content, which is often restricted by paywalls, copyright limitations, or technical barriers. Second, many websites implement measures such as anti-scraping rules, human verification steps, re-directions, or lazy loading mechanisms that complicate automated content extraction. Without the use of advanced browser automation tools, scraping full articles can be unreliable and may require ongoing manual intervention to adapt to changing website structures.

Focusing on titles also aligns with the typical user experience, as the title is usually the first—and sometimes the only—piece of information a reader encounters. Therefore, analyzing titles provides insight into the initial sentiment and framing presented to the audience.

\begin{figure*}
    \centering
    \includegraphics[width=\linewidth]{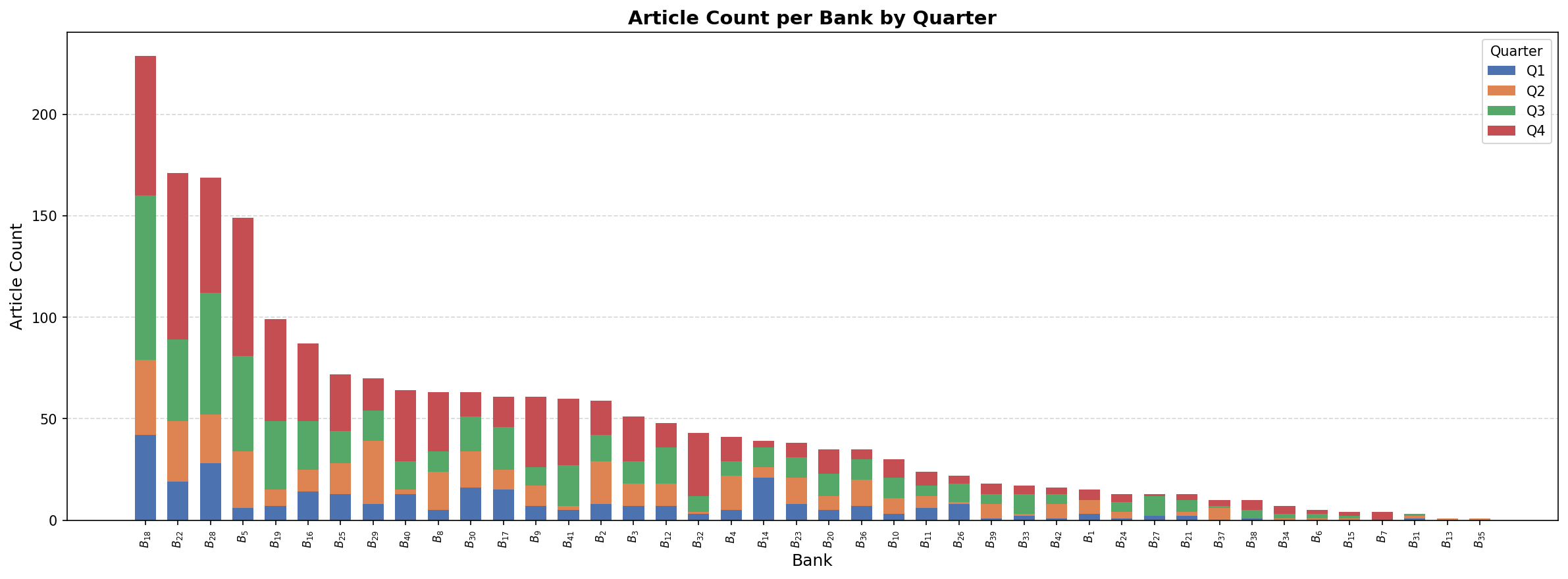}
    \label{img::term_per_firm}
    \caption{Quarterly article counts by bank in 2025. The bar chart reports the number of articles for each bank on the x-axis, with counts grouped by quarter (Q1--Q4).}
\end{figure*}

From the scraping phase, we collected a total of 2513 articles. Table~\ref{tab::terms_and_quarters} reports the quarterly distribution of articles in 2025 by category, while Figure~\ref{img::term_per_firm} shows the number of articles per company.

Table~\ref{tab::terms_and_quarters} shows that article counts vary substantially across both categories and time: several topic--firm combinations have zero observations in an entire quarter, while others cluster around specific events. This sparsity and temporal irregularity directly motivate the fixed-grid aggregation and causal filling stages. Within active periods, the same events are frequently reported across multiple outlets with only surface-form variation, motivating redundancy control. Finally, the distribution of classifier confidence (measured by entropy of $\boldsymbol{\pi}_{p,i}$) is heavy-tailed: a substantial fraction of titles produce near-uniform probability vectors, motivating uncertainty-aware weighting.
Similar variability is observed when articles are grouped by firm. Table \ref{img::term_per_firm} highlights marked disparities in the number of articles per company, reflecting heterogeneous levels of media coverage among banks. The quarterly breakdown further suggests that some banks receive higher media attention over time.

\section{Evaluation Design (Label-Free)}
\label{sec:evaluation_design}
Our evaluation is \emph{label-free}: we do not assume access to ground-truth sentiment labels, nor do we treat the sentiment classifier as an object of benchmarking. Instead, we evaluate the reconstructed sentiment time series as a \emph{signal reconstruction} output derived from sparse, redundant, and uncertain textual measurements.
All metrics are computed \emph{per bank} $p\in\mathcal{P}$, and then summarized \emph{across banks} using a fixed reporting block.

\subsection{Evaluation goals}
\label{sec:eval_goals}
A reconstruction pipeline is useful only if it (i)~reduces instability attributable to noisy/sparse measurements, (ii)~preserves information content (avoids excessive lag or amplitude collapse), (iii)~is robust to redundancy and measurement uncertainty, and (iv)~respects causality (no future leakage). Accordingly, our label-free evaluation covers:
\begin{enumerate}
    \item \textbf{Variance / stability control:} the reconstructed series should exhibit reduced high-frequency variability relative to raw aggregation.
    \item \textbf{Information preservation:} stability improvements should not come primarily from excessive lag or over-smoothing.
    \item \textbf{Robustness to redundancy and uncertainty:} repeated/near-duplicate stories and ambiguous titles should not disproportionately distort the signal.
    \item \textbf{Causality compliance:} the final output must be strictly causal; modifications at time $t_0$ must not affect values prior to $t_0$.
\end{enumerate}
Goals~(i) and~(ii) are evaluated via the metrics defined in Section~\ref{sec:eval_metrics}. Goals~(iii) and~(iv) are evaluated via the counterfactual tests in Section~\ref{sec:counterfactual_tests}.

We report results for any stage $S_{\mathrm{final}}^{(p)}\in\{S_A^{(p)}, S_F^{(p)}, S_S^{(p)}\}$, respecting the operator order
$S_A^{(p)}\leftarrow S^{(p)}$, $S_F^{(p)}\leftarrow S_A^{(p)}$, $S_S^{(p)}\leftarrow S_F^{(p)}$.

\subsection{Metrics (per bank)}
\label{sec:eval_metrics}
Let $X^{(p)}(k)$ denote a regular-grid series for bank $p$ on bins $k=1,\dots,T_p$ (e.g., weekly or monthly grid). Unless stated otherwise, metrics are computed on overlapping, non-missing indices for the relevant pair of series.

\subsubsection{Stability via total variation}
\label{sec:tv}
As a primary stability measure, we use total variation (TV), which quantifies the cumulative absolute change of a discrete-time signal:
\begin{equation}
\label{eq:tv_def}
\mathrm{TV}(X^{(p)})
=
\sum_{k=2}^{T_p} \left|X^{(p)}(k)-X^{(p)}(k-1)\right|.
\end{equation}
We compute TV for each reconstruction stage:
\begin{equation}
\label{eq:tv_stages}
\mathrm{TV}\!\left(S_A^{(p)}\right),\qquad
\mathrm{TV}\!\left(S_F^{(p)}\right),\qquad
\mathrm{TV}\!\left(S_S^{(p)}\right),
\end{equation}
and the corresponding relative reductions as ratios:
\begin{equation}
\label{eq:tv_ratios}
\frac{\mathrm{TV}(S_F^{(p)})}{\mathrm{TV}(S_A^{(p)})},\qquad
\frac{\mathrm{TV}(S_S^{(p)})}{\mathrm{TV}(S_F^{(p)})},\qquad
\frac{\mathrm{TV}(S_S^{(p)})}{\mathrm{TV}(S_A^{(p)})}.
\end{equation}
Lower TV indicates a less oscillatory series; ratios below $1$ indicate net smoothing (stability gain) from the corresponding operator.

\subsubsection{Gap behavior diagnostic for filling}
\label{sec:gap_drift}
To characterize how strongly the causal fill/projection deviates from a purely hold-based baseline \emph{only within missing segments}, we define a gap drift metric.
Let $M^{(p)}\subseteq \{1,\dots,T_p\}$ be the set of indices where the aggregated grid series is missing:
\[
M^{(p)} = \{\, k : S_A^{(p)}(k)=\mathrm{NA}\,\}.
\]
Inside each missing segment, define the hold reference $\widetilde S_A^{(p)}(k)$ as the last observed value before the gap (a causal baseline):
\begin{equation}
\label{eq:gap_hold_reference}
\widetilde S_A^{(p)}(k) = S_A^{(p)}(k^\star(k)),
\end{equation}
where
\begin{equation}
k^\star(k)=\max\{\,u<k : S_A^{(p)}(u)\neq \mathrm{NA}\,\}.
\end{equation}
The gap drift is then
\begin{equation} 
\label{eq:gap_drift}
\mathrm{GD}^{(p)}
=
\frac{1}{|M^{(p)}|}
\sum_{k\in M^{(p)}}
\left|S_F^{(p)}(k)-\widetilde S_A^{(p)}(k)\right|
\end{equation}
where $|M^{(p)}|>0.$
If $|M^{(p)}|=0$ (no missing bins for bank $p$), the gap drift is undefined and the bank is excluded from the cross-sectional summary of this metric.

This diagnostic measures the typical magnitude by which the chosen fill method departs from a causal hold reference, \emph{restricted to the bins that were missing in $S_A^{(p)}$}. By construction, $\mathrm{GD}^{(p)}=0$ for pure hold fill, and increases as the fill becomes more aggressive (e.g., decays toward a baseline).

\subsubsection{Information preservation via lag proxy}
\label{sec:lag_proxy}
Smoothing can introduce lag. To quantify lag between the filled series $S_F^{(p)}$ and the smoothed series $S_S^{(p)}$, we compute a small-lag cross-correlation proxy.
For lags $\ell\in\{-2,-1,0,1,2\}$, let $\rho^{(p)}(\ell)$ denote the Pearson correlation between $S_F^{(p)}(k)$ and $S_S^{(p)}(k+\ell)$ on the set of overlapping indices where both are observed.
We define
\begin{equation}
\label{eq:lag_star}
\ell^{*(p)}=\arg\max_{\ell\in\{-2,-1,0,1,2\}} \rho^{(p)}(\ell),
\end{equation}
and report per bank:
\begin{equation}
\label{eq:lag_outputs}
\rho^{(p)}(\ell^{*(p)}),
\qquad
|\ell^{*(p)}|.
\end{equation}
The peak correlation $\rho^{(p)}(\ell^{*(p)})$ captures alignment under an optimal small shift, while $|\ell^{*(p)}|$ provides a coarse lag magnitude proxy in grid units. Larger $|\ell^{*(p)}|$ indicates greater temporal displacement introduced by smoothing.

\subsection{Cross-sectional reporting across banks}
\label{sec:reporting}
For each metric $m^{(p)}$ computed per bank $p\in\mathcal{P}$, we report only cross-sectional summaries across banks (no per-bank tables in the main paper). Specifically, for each metric family we report:
\begin{equation}
\label{eq:reporting_block}
\mathrm{median}\ \pm\ \mathrm{IQR},
\qquad
\mathrm{mean}\ \pm\ \mathrm{std},
\qquad
q_{10},\ q_{90},
\end{equation}
where $q_{10}$ and $q_{90}$ are the 10th and 90th percentiles across the set $\{m^{(p)}\}_{p\in\mathcal{P}}$.
This fixed block provides a robust central tendency (median/IQR), a parametric summary (mean/std), and tail sensitivity ($q_{10}$/$q_{90}$) for bank heterogeneity.

\section{Counterfactual Tests}
\label{sec:counterfactual_tests}

The two tests in this section directly validate the two most important
engineering properties of the pipeline:
\begin{itemize}
  \item The \emph{impulse causality test} confirms that the pipeline is strictly causal and therefore deployable in a live, forward-looking setting without information leakage.
  \item The \emph{duplicate robustness test} confirms that redundancy control has a material practical effect, not only theoretical appeal.
\end{itemize}
Both tests are run on every evaluated pipeline configuration; only
inputs are modified, all operators and hyperparameters are held fixed.

\subsection{Impulse / causality test}
\label{sec:impulse_test}
A strictly causal pipeline must not transmit information backward in time: modifying observations at a time $t_0$ must not affect any reconstructed values strictly prior to $t_0$.
We test this property by injecting a localized impulse at time $t_0$ into $S^{(p)}$ and measuring any pre-$t_0$ deviations in the recomputed output.

\paragraph{Perturbation.}
For a given bank $p$, choose a grid index $k_0$ corresponding to time $t_0$ on the aggregation grid. Construct a perturbed sparse set $S_{\Delta}^{(p)}$ by either:
\begin{enumerate}
    \item \textbf{Additive impulse on an existing title:} select an index $i_0$ with $t_{p,i_0}\in D_{k_0}$ and set
    \[
    s^{\Delta}_{p,i_0}=s_{p,i_0}+\Delta,
    \qquad
    s^{\Delta}_{p,i}=s_{p,i}\ \ (i\neq i_0),
    \]
    with $S_{\Delta}^{(p)}=\{(t_{p,i}, s^\Delta_{p,i})\}_i$; or
    \item \textbf{Synthetic impulse observation:} add a new synthetic observation at time $t_0$,
    \[
    S_{\Delta}^{(p)} = S^{(p)} \cup \{(t_0,\Delta)\},
    \]
    representing a single additional measurement of magnitude $\Delta$.
\end{enumerate}
In either case, $\Delta$ is chosen such that the perturbation is clearly detectable in the local neighborhood of $k_0$ while remaining within the sentiment scale (e.g., clipping if needed to preserve $[-1,1]$). The exact injection scheme is fixed across all banks for comparability.

\paragraph{Re-run and outputs.}
Run the pipeline on both the original and perturbed inputs, producing (for a fixed chosen output stage)
\[
X^{(p)} = S_{\mathrm{final}}^{(p)}\bigl(S^{(p)}\bigr),
\qquad
X_\Delta^{(p)} = S_{\mathrm{final}}^{(p)}\bigl(S_\Delta^{(p)}\bigr),
\]
where $S_{\mathrm{final}}^{(p)}(\cdot)$ denotes the full pipeline applied to a given input observation set (overloading notation from Section~\ref{sec:sentiment_modeling}, where $S_{\mathrm{final}}^{(p)}$ denotes the output series itself).

\paragraph{Pre-$k_0$ deviation metric.}
Define the maximum pre-$k_0$ deviation
\begin{equation}
\label{eq:d_pre}
D_{\mathrm{pre}}^{(p)}
=
\max_{k<k_0}
\left|X_\Delta^{(p)}(k)-X^{(p)}(k)\right|.
\end{equation}
For a strictly causal reconstruction, $D_{\mathrm{pre}}^{(p)}$ should be zero up to numerical tolerance. We summarize $\{D_{\mathrm{pre}}^{(p)}\}_{p\in\mathcal{P}}$ across banks using the fixed reporting block (median/IQR, mean$\pm$std, $q_{10}/q_{90}$).
Optionally, we also report a pass rate under tolerance $\tau$:
\[
\mathrm{PassRate}(\tau)=\frac{1}{|\mathcal{P}|}\sum_{p\in\mathcal{P}} \mathbb{1}\!\left[D_{\mathrm{pre}}^{(p)}\le \tau\right],
\]
but magnitude summaries are always included.

\subsection{Duplicate robustness test}
\label{sec:dup_robustness_test}
This test isolates the effect of near-duplicate detection within aggregation by injecting synthetic near-duplicate titles into the raw sparse input and comparing reconstructions with redundancy control enabled versus disabled.

\paragraph{Baseline.}
For each bank $p$, compute the baseline smoothed series from the original sparse set:
\[
S_S^{(p)} = \mathrm{Pipeline}\bigl(S^{(p)}\bigr),
\]
where the pipeline configuration (aggregation grid, uncertainty weighting, fill method, filter method, and all hyperparameters) is fixed.

\paragraph{Constructing an augmented duplicate set.}
Construct
\[
S_{\mathrm{dup}}^{(p)} = S^{(p)} \cup \Delta S^{(p)},
\]
where $\Delta S^{(p)}$ is a set of injected synthetic near-duplicate titles designed to mimic real redundancy phenomena (syndication/paraphrase).
Concretely, for selected original titles in $S^{(p)}$, we generate paraphrases (e.g., via rule-based surface-form variation or LLM-based rephrasing) that preserve the underlying event semantics while varying surface form, and assign them timestamps within the same aggregation bin as their source titles so that they would be aggregated together absent detection. The specific generation method should be documented per experiment, as the quality and diversity of synthetic duplicates directly affects the test's validity.

\paragraph{Two matched configurations.}
Run two reconstructions on $S_{\mathrm{dup}}^{(p)}$, identical in all respects except redundancy control inside aggregation:
\begin{itemize}
    \item $S_{S,\mathrm{no\_detect}}^{(p)}$: duplicate detection \emph{OFF} (no redundancy correction in aggregation),
    \item $S_{S,\mathrm{detect}}^{(p)}$: duplicate detection \emph{ON} (redundancy correction enabled).
\end{itemize}
All other components (uncertainty weights, fill, and causal filter) remain unchanged.

\paragraph{Distance-to-baseline metrics.}
We quantify duplicate-induced distortion by $\ell_1$ distances to the baseline $S_S^{(p)}$:
\begin{equation}
\label{eq:dup_distances}
D_{\mathrm{no\_detect}}^{(p)}
=
\sum_{k}
\left|S_S^{(p)}(k)-S_{S,\mathrm{no\_detect}}^{(p)}(k)\right|,
\end{equation}
and
\begin{equation}\label{eq:dup_distances_2}
D_{\mathrm{detect}}^{(p)}
=
\sum_{k}
\left|S_S^{(p)}(k)-S_{S,\mathrm{detect}}^{(p)}(k)\right|.
\end{equation}
We report only cross-sectional summaries across banks (median/IQR, mean$\pm$std, $q_{10}/q_{90}$) for both $\{D_{\mathrm{no\_detect}}^{(p)}\}_p$ and $\{D_{\mathrm{detect}}^{(p)}\}_p$.
This test has \emph{no pass/fail criterion}: its purpose is to measure how much redundancy control mitigates duplicate-induced distortion in practice.

\paragraph{Interpretation.}
A successful redundancy-control mechanism should yield $D_{\mathrm{detect}}^{(p)} \ll D_{\mathrm{no\_detect}}^{(p)}$ for most banks, indicating that the reconstructed signal is substantially less sensitive to redundant observations when detection is enabled.

\section{Sentiment Signal Consistency Check}
\label{sec:experiment}

The sentiment modeling pipeline described in Section~\ref{sec:sentiment_modeling} produces, for each bank $p\in\mathcal{P}$, a reconstructed latent sentiment series $S_{\mathrm{final}}^{(p)}$ on a regular grid $\{D_1,\dots,D_T\}$. Before deploying such a signal in any downstream application, it is natural to ask whether it is \emph{plausible}: does it behave in a manner consistent with observable market data, and is its relationship to price stable across pipeline configurations?

We address this through a multi-method consistency study against bank stock prices. The study is not intended as a predictive benchmark. AI-related news sentiment is one of many factors influencing bank stock prices, most of which are unrelated to AI; for large, diversified institutions in particular, the marginal impact of AI sentiment on price is expected to be small. The results are therefore interpreted as a \emph{plausibility check} on signal behaviour rather than as evidence of predictive power. The inherent limitations of price correlation as a validation target are discussed in Section~\ref{sec:limitations}.

\subsection{Price series and transformations}
\label{sec:price_series}

For each bank $p$, let $P^{(p)}(k)$ denote the closing price at the end of aggregation period $D_k$, aligned to the same weekly grid as $S_{\mathrm{final}}^{(p)}$. Raw price levels are non-stationary and unsuited to correlation analysis alongside a bounded sentiment signal; we therefore work with return-based transformations throughout. Two representations are considered:

\begin{itemize}
    \item \textbf{Rolling min-max scaling} (window size $W$):
    Let
    $X_{\min}^{(p)}(k) = \min_{j \in \{k-W+1,\dots,k\}} X^{(p)}(j)$
    and
    $X_{\max}^{(p)}(k) = \max_{j \in \{k-W+1,\dots,k\}} X^{(p)}(j)$
    denote the trailing-$W$ minimum and maximum of $X^{(p)}$. Then
    \begin{equation}
        \label{eq:rolling_min_max}
        \widetilde{X}^{(p)}(k)
        =
        \frac{
            X^{(p)}(k) - X_{\min}^{(p)}(k)
        }{
            X_{\max}^{(p)}(k) - X_{\min}^{(p)}(k)
        },
    \end{equation}
    which rescales $X^{(p)}(k)$ to $[0,1]$ relative to its trailing-$W$ local range.

    \item \textbf{Rolling log return} ($\mathrm{RollRet}_W$):
    \begin{equation}
    \label{eq:roll_return}
    R_{W}^{(p)}(k) = \log P^{(p)}(k) - \log P^{(p)}(k - W),
    \end{equation}
    where $W$ matches the sentiment aggregation window. This representation integrates price movement over the same horizon as the sentiment signal, making the two series semantically comparable: both express accumulated directional movement over the same lookback period.
\end{itemize}

We generically write $Y^{(p)}(k)$ for the chosen representation. An
Augmented Dickey--Fuller (ADF) test is applied as a diagnostic safeguard~\cite{dickey1979distribution,dickey1981likelihood};
if residual non-stationarity is detected, the affected series is differenced
before use. In practice this rarely triggers; both return representations
are approximately stationary.

\subsection{Analysis notation}
\label{sec:notation_methods}

The following notation is used throughout this section. Let $S_{\mathrm{stat}}^{(p)}$ and $Y_{\mathrm{stat}}^{(p)}$ denote the stationary analysis versions of the sentiment and price series respectively: $S_{\mathrm{final}}^{(p)}$ after any required stationarity adjustment, and $Y^{(p)}$ as defined above after any analogous adjustment. Let $\tilde{S}^{(p)}$ and $\tilde{Y}^{(p)}$ denote the prewhitened residual series produced specifically for the CCF analysis in Section~\ref{sec:ccf}. The Granger and spectral analyses in Section~\ref{sec:granger_spectral} operate on $S_{\mathrm{stat}}^{(p)}$ and $Y_{\mathrm{stat}}^{(p)}$ directly, not on the CCF residuals.

\subsection{Primary methods}
\label{sec:correlation_methods}

We employ two primary methods targeting distinct aspects of the sentiment--price relationship: static lead/lag structure net of autocorrelation, and directional co-movement with frequency decomposition. A supplementary exploratory diagnostic is described in Section~\ref{sec:dtw}.

\subsubsection{Prewhitened cross-correlation function}
\label{sec:ccf}

A direct cross-correlation between $S_{\mathrm{stat}}^{(p)}$ and $Y_{\mathrm{stat}}^{(p)}$ is confounded by the autocorrelation structure present in both series: any smooth signal will appear correlated with its own lagged version purely because of serial dependence, inflating apparent cross-correlations at multiple lags. We address this using a matched-order residualization procedure motivated by Box--Jenkins prewhitening~\cite{box1970distribution, box2013box}.

An $\mathrm{ARIMA}(p_0, 0, q_0)$ model~\cite{akaike1974aic} is fitted to $S_{\mathrm{stat}}^{(p)}$, with order $(p_0, q_0)$ selected by minimising the Akaike information criterion over the grid $p_0 \in \{0,\dots,4\}$, $q_0 \in \{0,1,2\}$, yielding residuals $\tilde{S}^{(p)}$. The same order \emph{structure} $(p_0, q_0)$ is then used to fit a separate $\mathrm{ARIMA}$ model to $Y_{\mathrm{stat}}^{(p)}$, yielding residuals $\tilde{Y}^{(p)}$.

This is not the strict Box--Jenkins coefficient-transfer procedure, in which the estimated filter from the sentiment series is applied directly to the price series without refitting. Instead, it is a practical applied approximation that removes autocorrelation of the same order class from both series while avoiding coefficient-transfer instability in short samples and mixed ARMA specifications. For the present purpose---a plausibility-oriented check of lead/lag structure rather than structural transfer-function identification---this approximation is adequate and materially reduces spurious cross-lag dependence.

For a given lag $\kappa$ (in weeks), the prewhitened cross-correlation is:
\begin{equation}
\label{eq:pwccf}
\hat{\rho}^{(p)}(\kappa)
= \frac{\sum_{k} \tilde{S}^{(p)}(k)\,\tilde{Y}^{(p)}(k+\kappa)}
       {\sqrt{\sum_{k}\tilde{S}^{(p)}(k)^2}\,\sqrt{\sum_{k}\tilde{Y}^{(p)}(k)^2}},
\end{equation}
evaluated over $\kappa \in \{-\kappa_{\max},\dots,\kappa_{\max}\}$ with $\kappa_{\max} = 12$ weeks. Positive $\kappa$ corresponds to sentiment leading price. As an approximate reference under the white-noise null, a 95\% confidence band of $\pm 1.96/\sqrt{n}$ is reported, where $n$ is the number of matched residual pairs. Given the approximate nature of the residualization step, these bands are interpreted descriptively rather than as exact finite-sample inference.

\subsubsection{Rolling Granger causality and spectral coherence}
\label{sec:granger_spectral}

The CCF identifies association and lead/lag structure but does not establish directional predictability. We therefore complement it with two analyses operating on $S_{\mathrm{stat}}^{(p)}$ and $Y_{\mathrm{stat}}^{(p)}$ directly.

\paragraph{Rolling Granger causality.}
For each bank $p$, a bivariate VAR is fitted to $(S_{\mathrm{stat}}^{(p)},\, Y_{\mathrm{stat}}^{(p)})$ and we test whether past values of $S_{\mathrm{stat}}^{(p)}$ improve one-step-ahead prediction of $Y_{\mathrm{stat}}^{(p)}$ beyond what past values of $Y_{\mathrm{stat}}^{(p)}$ alone provide, and vice versa~\cite{granger1969causality}. The lag order $\ell$ is selected once on the full sample via the VAR AIC criterion and held fixed across all rolling windows of size $W_{\mathrm{roll}}$ to ensure comparability. Windows containing fewer than $\max(0.8\,W_{\mathrm{roll}},\; 4\ell + 10)$ valid observations are excluded to avoid unstable $F$-statistics in short effective samples.

The test statistic is the SSR-based $F$-statistic from the restricted versus unrestricted VAR, evaluated at each window endpoint $D_k$. We report the direction of any significant ($p < 0.05$) Granger causality per window, and summarise across windows by the net log $F$-ratio:
\begin{equation}
\label{eq:net_granger}
\Delta_G^{(p)}(k)
= \log F_{S_{\mathrm{stat}}\to Y_{\mathrm{stat}}}(k)
- \log F_{Y_{\mathrm{stat}}\to S_{\mathrm{stat}}}(k),
\end{equation}
where $\Delta_G^{(p)}(k) > 0$ indicates that sentiment more strongly Granger-causes price than the reverse. Granger causality establishes incremental linear predictability within the VAR~\cite{granger1969causality, sims1980macroeconomics}, not structural causation; a significant result is consistent with both genuine sentiment-driven price movement and unmodelled common factors driving both series.

\paragraph{Spectral coherence and phase.}
To decompose the sentiment--price relationship by frequency, we compute the squared coherence between $S_{\mathrm{stat}}^{(p)}$ and $Y_{\mathrm{stat}}^{(p)}$ using Welch's method with segment length $n_{\mathrm{seg}}$ weeks~\cite{welch1967fft}. Squared coherence at frequency $f$ is a normalised measure of linear association between the two series at that frequency, independent of scale.

The cross-spectral phase angle $\phi(f)$ is used to estimate the apparent lead of sentiment over price:
\begin{equation}
\label{eq:phase_lag}
\tau(f) = \frac{\phi(f)}{2\pi f} \quad \text{(weeks)},
\end{equation}
where $\tau(f) > 0$ indicates that $S_{\mathrm{stat}}^{(p)}$ leads $Y_{\mathrm{stat}}^{(p)}$. This estimate is unambiguous provided $|\tau(f)| < 1/(2f)$. For the expected sentiment--price lags of one to four weeks and the frequency bands analysed here, this condition is satisfied. Phase estimates are reported only in frequency bands where coherence is non-negligibly above zero; in bands with near-zero coherence, phase angles are dominated by noise and are not interpreted. Results are summarised by frequency band: short-run ($\leq 4$ weeks), medium-run ($4$--$13$ weeks), and long-run ($>13$ weeks). Long-run band estimates are the least stable because they are based on relatively few effective cycles in the available sample and are therefore treated as descriptive rather than inferential.

For pipeline comparison, we define the \emph{medium-run coherence fraction} for bank $p$ as
\begin{equation}
\label{eq:medium_coherence_fraction}
C_{\mathrm{mid}}^{(p)}
=
\frac{\displaystyle \int_{f \in \mathcal{B}_{\mathrm{mid}}} \operatorname{Coh}^{(p)}(f)\,df}
     {\displaystyle \int_{f \in \mathcal{B}_{\mathrm{short}} \cup \mathcal{B}_{\mathrm{mid}} \cup \mathcal{B}_{\mathrm{long}}} \operatorname{Coh}^{(p)}(f)\,df},
\end{equation}
where $\operatorname{Coh}^{(p)}(f)$ denotes squared coherence and the three bands correspond respectively to short-, medium-, and long-run horizons defined above. In implementation, the integrals are replaced by sums over the discrete frequency grid returned by Welch's method. This quantity measures the share of non-trend coherence concentrated in the medium-run band.

\subsection{Supplementary diagnostic: rolling DTW-Pearson}
\label{sec:dtw}

As a supplementary shape-alignment diagnostic, we compute a time-varying co-movement measure using dynamic time warping (DTW). This method is not treated as primary evidence; its outputs are reported alongside the primary methods as a robustness lens on local elastic alignment and do not carry independent inferential weight. DTW-based aligned-path correlation is more flexible than classical CCF or VAR-based methods, but also harder to interpret inferentially, so its role in the comparison is explicitly subordinate to the CCF and Granger results.

The motivation is that the lead/lag structure between sentiment and price may vary over time. Within a rolling window of $W_{\mathrm{roll}}$ weeks ending at $D_k$, both $S_{\mathrm{stat}}^{(p)}$ and $Y_{\mathrm{stat}}^{(p)}$ are standardised to zero mean and unit variance. DTW alignment is then computed under a Sakoe--Chiba band constraint~\cite{sakoe2003dynamic} of $b$ weeks, limiting the maximum permissible alignment shift between matched points:
\begin{equation}
\label{eq:dtw_pearson}
r_{\mathrm{DTW}}^{(p)}(k)
= \mathrm{Pearson}\!\bigl(\mathbf{x}^*,\, \mathbf{y}^*\bigr),
\end{equation}
where $(\mathbf{x}^*, \mathbf{y}^*)$ denotes the aligned path vectors returned by the DTW algorithm. The band $b$ is set equal to $\kappa_{\max}$ so that the maximum alignment shift is comparable to the lag range used in the CCF analysis.

Since DTW alignment introduces repeated path points by construction, aligned vectors are not iid and standard parametric $p$-values are anti-conservative. Significance is therefore assessed via a block-permutation test: the price series within each window is permuted in blocks of size $b$, preserving local autocorrelation structure, and the $p$-value is the fraction of permuted statistics exceeding $|r_{\mathrm{DTW}}^{(p)}(k)|$ in absolute value. The block size is a practical calibration choice intended to preserve dependence over approximately the same local horizon as the alignment tolerance; it is not theoretically implied by the DTW band constraint.

\subsection{Cross-bank aggregation}
\label{sec:cross_bank}

All quantities are computed independently for each bank $p \in \mathcal{P}^*$, where $\mathcal{P}^* \subseteq \mathcal{P}$ denotes the set of banks with sufficient matched observations.

For scalar correlation quantities (prewhitened CCF at a given lag $\kappa$; mean DTW-Pearson as a secondary summary), cross-bank aggregation is performed using Fisher's $r$-to-$z$ transformation~\cite{fisher1921probable,fisher1915frequency}:
\begin{equation}
\label{eq:fisher_transform}
z^{(p)} = \operatorname{arctanh}\!\bigl(\rho^{(p)}\bigr),
\end{equation}
which maps $\rho \in (-1,1)$ to an approximately normally distributed variable. Three aggregation schemes are considered:
\begin{align}
\label{eq:fisher_agg}
\bar{z}^{\mathrm{mean}} &= \frac{1}{|\mathcal{P}^*|}\sum_{p\in\mathcal{P}^*} z^{(p)},
\\[4pt]
\bar{z}^{\mathrm{wt}} &= \frac{\sum_{p\in\mathcal{P}^*} w_p\, z^{(p)}}{\sum_{p\in\mathcal{P}^*} w_p},\\
&w_p \in \bigl\{N_p,\,\sqrt{N_p},\,\log(1+N_p)\bigr\},
\\[4pt]
\bar{z}^{\mathrm{med}} &= \operatorname{median}_{p\in\mathcal{P}^*}\bigl\{z^{(p)}\bigr\},
\end{align}
where $N_p = \sum_k t_{p,k}$ is the total article count for bank $p$, with $t_{p,k} = |I_{p,k}|$ as defined in Section~\ref{sec:smoothing}. Linear weighting gives the most-covered banks the greatest influence; square-root and log weightings progressively compress this dynamic range. The median serves as a robust alternative insensitive to extreme per-bank values. The aggregate correlation is recovered as $\bar{\rho} = \tanh(\bar{z})$.

Cross-bank significance for the CCF is assessed using Fisher's combined probability test~\cite{fisher1932statistical}: under the null of zero correlation at lag $\kappa$, the statistic $-2\sum_{p}\log p^{(p)}(\kappa)$ follows a $\chi^2$ distribution with $2|\mathcal{P}^*|$ degrees of freedom. This test is used here as a heuristic meta-analytic summary. Its formal validity requires approximate independence of the per-bank test statistics after prewhitening; however, banks in the same sector share macro exposures, rate environments, and market-regime effects that can induce non-trivial cross-bank dependence even after within-series filtering. Combined $p$-values are therefore treated as approximate and interpreted with corresponding caution rather than as precise significance statements.

For Granger and spectral quantities, cross-bank summaries are reported as means and standard deviations across $\mathcal{P}^*$; no Fisher transform is applied since these quantities are not correlation coefficients.

\subsection{Pipeline configurations}
\label{sec:pipeline_configs}

The consistency check is run independently for each pipeline configuration under evaluation. Each configuration corresponds to a specific combination of choices within the framework of Section~\ref{sec:sentiment_modeling}: the uncertainty weighting scheme $W_{\mathrm{unc}}$, the redundancy control method $\phi$, the filling function $g$, and the smoothing operator $\mathcal{F}$. All configurations use global aggregation and weekly bins (W-FRI grid); the aggregation frequency is held fixed across configurations so that differences in consistency estimates are attributable to pipeline design rather than temporal resolution.

For each configuration, the methods above produce the following scalar summaries, constituting a row of the cross-strategy comparison table: peak prewhitened CCF aggregate $\bar{\rho}$ and its lag $\kappa^*$; percentage of rolling windows with significant Granger causality in the sentiment-to-price direction; and aggregated medium-run coherence fraction based on Eq.~\eqref{eq:medium_coherence_fraction}. Mean rolling DTW-Pearson is additionally reported as a supplementary column. The full set of evaluated configurations and the resulting comparison are presented in Section~\ref{sec:results}.

\subsection{Interpretation of Results}

The results of the sentiment–price consistency analysis should be
interpreted with several caveats in mind.

First, the sentiment signal analyzed in this study is derived from
AI-related news coverage. While such information may influence market
perceptions, it represents only a small component of the information set affecting the valuation of large financial institutions. Macroeconomic conditions, interest rates, regulatory developments, and firm-specific events typically play a much larger role in determining stock price movements. \par
Consequently, the relationship between the reconstructed sentiment signal and price dynamics is expected to be modest. Second, the statistical diagnostics used in this analysis rely on standard time-series assumptions that may hold only approximately in practice. For example, the Augmented Dickey–Fuller test used to assess stationarity has limited power in relatively short samples~\cite{elliott1992efficient}. In this study, stationarity is primarily addressed through return-based transformations, while the ADF test serves as a diagnostic safeguard rather than the primary adjustment mechanism. \par
Third, the number of banks included in the cross-sectional analysis is
limited. Individual firm-level results may therefore be noisy, and
cross-bank aggregates should be interpreted as descriptive summaries
rather than precise statistical estimates.
Fourth, Granger causality tests identify incremental predictive
relationships within the specified vector autoregression model but do
not establish structural causation. A statistically significant result
simply indicates that past sentiment information improves prediction of price movements within the model. \par
Finally, several aggregation procedures rely on approximate statistical assumptions. In particular, combined significance measures assume independence across banks, whereas financial institutions often share common macroeconomic exposures and market regimes. As a result, combined p-values should be interpreted as heuristic indicators rather than exact statistical inference.
Taken together, these considerations suggest that the consistency
analysis should be viewed primarily as a plausibility check for the
reconstructed sentiment signal rather than as evidence of strong
predictive power.

\section{Results}
\label{sec:results}

This section follows the validation hierarchy stated in Section~\ref{sec:intro}. We begin by showing that the raw article stream has the exact pathologies that motivate reconstruction: irregular coverage and an aggregation trade-off between sparsity reduction and semantic mixing. We then report two internal counterfactual checks that test whether the pipeline behaves as intended under controlled perturbations. Finally, we turn to the sequential design study, where candidate component choices are compared stage by stage and the retained global and local specifications are identified. The external market-consistency analysis is reported separately in Section~8.4.

\subsection{Data properties: why reconstruction is necessary}
\label{sec:data_properties}

Before comparing reconstruction variants, it is necessary to establish that the raw article stream is not already suitable for direct use. Two diagnostics matter here. First, the observation process may be too sparse and too irregular to support a stable firm-level signal on a regular grid without an explicit projection rule. Second, even if one coarsens the grid to reduce missingness, the resulting bins may become semantically heterogeneous, weakening the interpretation of the aggregated signal.

Table~\ref{tab::data_properties_combined} brings these two diagnostics together. Panel~A reports coverage and inter-arrival properties of the raw stream. Cross-sectional article counts differ markedly across firms, while the inter-arrival distribution shows that multi-day publication gaps are routine rather than exceptional. The median gap is already 2.5 days, and the upper quartile reaches 8 days. Missing intervals are therefore a structural feature of the data-generating process, not a peripheral inconvenience. Any reconstruction on a fixed temporal grid must consequently decide how information is propagated through periods with no new articles.

Panel~B then quantifies the cost of reducing sparsity through coarser aggregation. Within-bin sentiment dispersion rises monotonically as one moves from daily to weekly, monthly, and quarterly bins. This is the central aggregation trade-off. Narrow bins preserve local semantic coherence but leave the signal too sparse for robust reconstruction; wide bins reduce missingness but pool increasingly heterogeneous news items and therefore blend distinct local sentiment states. Weekly aggregation remains the most credible operational compromise. It reduces sparsity substantially relative to the daily grid without introducing the much heavier semantic mixing visible at monthly and quarterly frequencies. For that reason, the main experiments are conducted on a weekly grid.

\begin{table*}[!ht]
    \centering
    \caption{Diagnostic motivation for reconstruction. Panel~A reports cross-sectional coverage irregularity in the raw article stream. Panel~B reports within-bin sentiment dispersion under alternative aggregation periods. Together, the two panels motivate a weekly reconstruction grid with explicit causal filling.}
    \label{tab::data_properties_combined}
    \begin{tabularx}{\textwidth}{@{}Xrrr@{}}
    \toprule
    \multicolumn{4}{@{}l@{}}{\textbf{Panel A: Coverage and inter-arrival diagnostics}}\\
    \midrule
    Statistic & median & Q1 & Q3 \\
    \midrule
    Article count & 46.0 & 13.5 & 77.0 \\
    Time gap & 2.5 days & 1 day & 8 days \\
    \midrule
    \multicolumn{4}{@{}l@{}}{\textbf{Panel B: Within-bin volatility by aggregation period}}\\
    \midrule
    Aggregation & median & Q1 & Q3 \\
    \midrule
    Daily (D) & 0.0243 & 7.6807e-8 & 0.1622 \\
    Weekly (W) & 0.1453 & 0.0758 & 0.3573 \\
    Monthly (ME) & 0.3191 & 0.1958 & 0.5268 \\
    Quarterly (QE) & 0.4774 & 0.3070 & 0.5444 \\
    \bottomrule
    \end{tabularx}
\end{table*}

\subsection{Counterfactual test results}
\label{sec:counterfactual_results}

Before optimizing component choices, the reconstruction pipeline must satisfy two non-negotiable engineering requirements. It must be strictly causal, so that no future information contaminates earlier timestamps, and it must be robust to redundant textual evidence, so that repeated near-duplicate articles do not spuriously amplify the signal. These are not ordinary performance comparisons. The first is a compliance check; the second is a robustness check.

Table~\ref{tab::counterfactual_combined} reports both diagnostics in a single two-panel format. Panel~A summarizes the impulse causality test. The result is exact: pre-impulse changes are zero throughout, and the pass rate is one across all reported summaries. This is the desired outcome. The table is sparse because the purpose of the test is to detect violations, and none are observed.

Panel~B reports the duplicate-injection experiment. The relevant comparison is between duplicate-unaware reconstruction (DD) and duplicate-aware reconstruction (DE), each examined before and after filtering. The pattern is clear. Activating duplicate control sharply reduces deviation from the unperturbed signal, whereas filtering produces only a secondary incremental gain once redundancy handling is already in place. Redundancy control is therefore not merely a conceptual safeguard; it has a material empirical effect on the reconstructed series and should be treated as a core design requirement rather than an optional refinement.

\begin{table*}[!ht]
    \centering
    \caption{Counterfactual validation of the reconstruction pipeline. Panel~A reports the impulse causality test, where zero pre-impulse change and unit pass rate indicate strict causality. Panel~B reports duplicate robustness under injected near-duplicate observations; lower values indicate smaller deviations from the unperturbed signal.}
    \label{tab::counterfactual_combined}
    \begin{tabularx}{\textwidth}{@{}Xrrrrrr@{}}
        \toprule
        \multicolumn{7}{@{}l@{}}{\textbf{Panel A: Impulse causality test}}\\
        \midrule
        Statistic & median & IQR & mean & std & p10 & p90 \\
        \midrule
        Max absolute pre-impulse change & 0.0 & 0.0 & 0.0 & 0.0 & 0.0 & 0.0 \\
        Pass rate & 1.0 & 0.0 & 1.0 & 0.0 & 1.0 & 1.0 \\
        \midrule
        \multicolumn{7}{@{}l@{}}{\textbf{Panel B: Duplicate robustness test}}\\
        \midrule
        Setting & median & IQR & mean & std & p10 & p90 \\
        \midrule
        DD & 0.2253 & 0.4287 & 0.2649 & 0.2587 & 0.0 & 0.6190 \\
        DE & 0.0120 & 0.0653 & 0.0455 & 0.0638 & 0.0 & 0.1470 \\
        DD + filtering & 0.1519 & 0.3850 & 0.2370 & 0.2364 & 0.0 & 0.5512 \\
        DE + filtering & 0.0117 & 0.0604 & 0.0478 & 0.0912 & 0.0 & 0.1304 \\
        \bottomrule
    \end{tabularx}
\end{table*}

\subsection{Sequential design study}
\label{sec:design_study}

We now turn to design selection. The purpose of this section is not exhaustive hyperparameter tuning, but disciplined justification of the retained reconstruction pipeline. Candidate variants are evaluated sequentially by pipeline stage. Within each stage, comparisons are made against the current retained baseline, and the preferred option is then carried forward. To reduce fragmentation and make the cross-regime logic explicit, the results are organized by stage rather than by separate global and local narrative blocks.

\subsubsection{Aggregation-stage choices}

The first stage concerns how article-level evidence is pooled before filling and smoothing. Three component families are examined: uncertainty weighting, redundancy control, and time decay. Table~\ref{tab::aggregation_stage_combined} reports these comparisons jointly, with global aggregation in Panel~A and local aggregation in Panel~B. Throughout this subsection, negative values indicate lower total variation than the stage-specific baseline.

The evidence shows that aggregation-stage choices matter, but mostly at the margin. Under global aggregation, uncertainty weighting yields only modest changes relative to the baseline. Entropy weighting produces the largest median reduction in total variation, but its profile is less even across the remaining summaries. Polarity weighting delivers a smaller but more stable pattern of change and is therefore retained as the global uncertainty-weighting rule. Under local aggregation, by contrast, all uncertainty-weighting variants remain close to the baseline. None provides a sufficiently consistent gain to justify additional complexity, so the local specification is kept unweighted.

Redundancy control behaves differently across regimes. In the global case, neither deduplication nor corroboration alters stability dramatically, but corroboration is the least disruptive overall and remains substantively plausible because repeated cross-outlet coverage may proxy salience once all AI-related articles are pooled into a single firm-level stream. In the local case, thematic separation already narrows the semantic scope of each stream, so repeated coverage is less naturally interpreted as independent support and more naturally treated as repeated evidence. In that setting, deduplication is the cleaner retained choice.

Time decay also reveals a regime-dependent pattern. Under global aggregation, decay functions act mainly as soft recency corrections, not as dominant stabilizers; the $\tau=5$ specification provides the most balanced overall profile and is therefore retained. Under local aggregation, decay matters more. All tested rules reduce total variation relative to the baseline, and the $\lambda=0.95$ specification produces the strongest improvement across the reported summaries. This suggests that category-level local streams benefit more from emphasizing recent observations than the pooled global stream does.

Taken together, the aggregation-stage results establish two points. First, early-stage design choices are empirically relevant, but they do not dominate the stability properties of the final signal. Second, the preferred settings are not identical across regimes. The retained aggregation-stage configuration is therefore polarity weighting, corroboration, and $\tau=5$ decay for the global pipeline, versus no uncertainty weighting, deduplication, and $\lambda=0.95$ decay for the local pipeline.

\begin{table*}[!ht]
    \centering
    \caption{Aggregation-stage comparisons, reported as differences in $\mathrm{TV}\!\left(S_A\right)$ relative to the stage-specific baseline. Negative values indicate lower total variation than the baseline. Panel~A reports the global regime; Panel~B reports the local regime.}
    \label{tab::aggregation_stage_combined}
    \begin{tabular}{llrrrrrr}
        \toprule
        \multicolumn{8}{l}{\textbf{Panel A: Global aggregation}}\\
        \midrule
        Component & Method & $\Delta$ median & $\Delta$ IQR & $\Delta$ mean & $\Delta$ std & $\Delta p_{10}$ & $\Delta p_{90}$ \\
        \midrule
        Uncertainty & Base & 8.1641 & 10.4489 & 10.1980 & 7.9933 & 0.9690 & 20.8326 \\
        Uncertainty & Entropy & -0.1888 & 1.0980 & 0.3221 & 0.6463 & -0.1721 & 1.5669 \\
        Uncertainty & Polarity & -0.0458 & 0.0090 & -0.0756 & 0.0066 & -0.1563 & 0.2876 \\
        Uncertainty & Top 2 Margin & 0.0522 & 2.8097 & 0.6793 & 1.1128 & -0.1332 & 2.2594 \\
        \addlinespace
        Redundancy & Base & 8.1184 & 10.4579 & 10.1224 & 7.9999 & 0.8127 & 21.1202 \\
        Redundancy & Deduplication & -0.1062 & 0.2524 & 0.1045 & 0.1878 & 0.1329 & -0.2563 \\
        Redundancy & Corroboration & 0.0131 & 0.0437 & -0.0147 & -0.0379 & 0.0 & -0.0902 \\
        \addlinespace
        Time decay & Base & 8.1315 & 10.5016 & 10.1077 & 7.9620 & 0.8127 & 21.0300 \\
        Time decay & Exp. decay ($\lambda=0.95$) & 1.0829 & 1.3350 & 0.3002 & -0.9230 & 0.3978 & -1.4903 \\
        Time decay & Exp. decay ($\tau=5$) & 0.3067 & -0.2542 & -0.1100 & -0.3037 & 0.1633 & -0.3556 \\
        Time decay & Exp. decay ($\alpha=0.95$) & 0.9605 & -0.0128 & -0.1216 & -0.7190 & 0.2356 & -0.9979 \\
        \midrule
        \multicolumn{8}{l}{\textbf{Panel B: Local aggregation}}\\
        \midrule
        Component & Method & $\Delta$ median & $\Delta$ IQR & $\Delta$ mean & $\Delta$ std & $\Delta p_{10}$ & $\Delta p_{90}$ \\
        \midrule
        Uncertainty & Base & 5.3482 & 7.2151 & 6.2272 & 5.4090 & 1.2085 & 11.3204 \\
        Uncertainty & Entropy & -0.0293 & 0.1137 & 0.3347 & 0.4895 & 0.0 & 1.4702 \\
        Uncertainty & Polarity & 0.0118 & -0.1690 & 0.0824 & 0.1406 & 0.0 & 0.5060 \\
        Uncertainty & Top 2 Margin & 0.0270 & 0.3433 & 0.4145 & 0.5916 & 0.0 & 1.3526 \\
        \addlinespace
        Redundancy & Base & 5.3482 & 7.2151 & 6.2272 & 5.4090 & 1.2085 & 11.3204 \\
        Redundancy & Deduplication & 0.0 & -0.0452 & -0.0175 & -0.0128 & 0.0 & -0.1556 \\
        Redundancy & Corroboration & 0.0 & -0.0206 & 0.0139 & 0.0049 & 0.0 & 0.0261 \\
        \addlinespace
        Time decay & Base & 5.3482 & 7.1699 & 6.2097 & 5.3962 & 1.2085 & 11.1648 \\
        Time decay & Exp. decay ($\lambda=0.95$) & -0.3482 & -1.0763 & -0.4654 & -0.6197 & 0.0 & -1.2546 \\
        Time decay & Exp. decay ($\alpha=5$) & -0.0886 & -0.3191 & -0.1263 & -0.1551 & 0.0 & -0.2248 \\
        Time decay & Exp. decay ($\tau=5$) & -0.2391 & -0.7934 & -0.3520 & -0.4261 & 0.0 & -0.6336 \\
        \bottomrule
    \end{tabular}
\end{table*}

\subsubsection{Filling-stage choices}

The second stage concerns how the signal is propagated across missing bins once the article stream has been projected onto a regular grid. Table~\ref{tab::filling_stage_combined} compares constant fill with constant-decay carry-forward rules under both aggregation regimes. The upper block in each panel reports the effect on post-fill total variation, while the lower block reports gap drift, that is, the extent to which a fill rule distorts the signal precisely inside intervals with no new publications.

The shared pattern across regimes is straightforward. Short-horizon decay rules are markedly more interventionist than simple carry-forward persistence: they increase instability and generate larger within-gap distortions. As the carry-forward horizon grows, the decayed rules approach ordinary persistence. The filling decision is therefore partly normative. Short horizons encode an aggressive prior that sentiment should revert quickly in the absence of fresh news, whereas long horizons represent only gradual attenuation.

The preferred rule nevertheless differs between regimes. In the global pipeline, constant fill remains the more conservative and more stable option. Since the objective is to recover a latent firm-level sentiment process under sparse observation, rather than to impose rapid mean reversion during silent periods, simple persistence is the better default. In the local pipeline, a long-horizon decayed carry-forward rule with $H=60$ provides a better compromise. It allows gradual attenuation in sparse category-level streams while avoiding the sharp instability associated with short-horizon decay.

The main substantive conclusion is therefore not that one universal fill rule dominates, but that both regimes favour conservative filling behaviour. What is rejected is not persistence itself, but aggressive short-horizon decay. Relative to the aggregation-stage results, the filling stage matters more for the local pipeline, yet it still does not dominate the final signal to the same degree as smoothing does.

\begin{table*}[!ht]
    \centering
    \caption{Filling-stage diagnostics under global and local aggregation. In each panel, the upper block reports differences in $\mathrm{TV}\!\left(S_F\right)$ relative to constant fill, while the lower block reports gap-drift values. Smaller values indicate less distortion.}
    \label{tab::filling_stage_combined}
    \begin{tabular}{llrrrrrr}
    \toprule
    \multicolumn{8}{l}{\textbf{Panel A: Global aggregation}}\\
    \midrule
    Method & Metric & $\Delta$ median & $\Delta$ IQR & $\Delta$ mean & $\Delta$ std & $\Delta p_{10}$ & $\Delta p_{90}$ \\
    \midrule
    Constant Fill & TV & 8.4382 & 10.2474 & 9.9976 & 7.6583 & 0.9759 & 20.6744 \\
    Constant Decay ($H=5$) & TV & 14.0754 & 10.2076 & 12.4371 & 5.1846 & 4.3472 & 20.5729 \\
    Constant Decay ($H=10$) & TV & 11.5373 & 6.7176 & 9.7749 & 3.0495 & 4.3472 & 11.8867 \\
    Constant Decay ($H=30$) & TV & 7.4817 & 0.7952 & 5.2929 & 0.4028 & 3.7988 & 3.4797 \\
    Constant Decay ($H=60$) & TV & 4.2455 & -0.4476 & 3.1054 & -0.2046 & 3.2627 & 1.8878 \\
    Constant Decay ($H=90$) & TV & 3.1539 & 0.2455 & 2.0596 & -0.2412 & 2.0743 & 1.3447 \\
    Constant Fill & Gap & 0.0 & 0.0 & 0.0 & 0.0 & 0.0 & 0.0 \\
    Constant Decay ($H=5$) & Gap & 0.5687 & 0.3181 & 0.5638 & 0.1909 & 0.3154 & 0.8248 \\
    Constant Decay ($H=10$) & Gap & 0.3699 & 0.2923 & 0.4256 & 0.2008 & 0.2134 & 0.7131 \\
    Constant Decay ($H=30$) & Gap & 0.1405 & 0.1877 & 0.2237 & 0.1848 & 0.0733 & 0.5215 \\
    Constant Decay ($H=60$) & Gap & 0.0703 & 0.0988 & 0.1337 & 0.1472 & 0.0366 & 0.3565 \\
    Constant Decay ($H=90$) & Gap & 0.0468 & 0.0664 & 0.0966 & 0.1257 & 0.0244 & 0.2377 \\
    \midrule
    \multicolumn{8}{l}{\textbf{Panel B: Local aggregation}}\\
    \midrule
    Method & Metric & $\Delta$ median & $\Delta$ IQR & $\Delta$ mean & $\Delta$ std & $\Delta p_{10}$ & $\Delta p_{90}$ \\
    \midrule
    Constant Fill & TV & 7.4912 & 8.5033 & 8.7847 & 7.4972 & 1.4911 & 16.1625 \\
    Constant Decay ($H=5$) & TV & 2.6540 & -0.5987 & 1.4389 & -1.2391 & 1.4083 & 0.1598 \\
    Constant Decay ($H=10$) & TV & 2.0282 & -1.0691 & 0.8656 & -1.7050 & 1.2238 & -1.2231 \\
    Constant Decay ($H=30$) & TV & 0.9188 & -1.5474 & -0.0563 & -2.0611 & 1.2165 & -2.6146 \\
    Constant Decay ($H=60$) & TV & 0.3416 & -1.4681 & -0.4278 & -1.8194 & 1.2381 & -2.5295 \\
    Constant Decay ($H=90$) & TV & 0.1428 & -1.4966 & -0.5942 & -1.5392 & 1.0158 & -2.7146 \\
    Constant Fill & Gap & 0.1254 & 0.0511 & 0.1211 & 0.0419 & 0.0713 & 0.1622 \\
    Constant Decay ($H=5$) & Gap & 0.0818 & 0.0694 & 0.1342 & 0.1083 & 0.0589 & 0.2837 \\
    Constant Decay ($H=10$) & Gap & 0.0504 & 0.0634 & 0.1014 & 0.1005 & 0.0384 & 0.1763 \\
    Constant Decay ($H=30$) & Gap & -0.0025 & 0.0374 & 0.0523 & 0.0829 & 0.0108 & 0.1300 \\
    Constant Decay ($H=60$) & Gap & -0.0189 & 0.0172 & 0.0267 & 0.0569 & 0.0132 & 0.1160 \\
    Constant Decay ($H=90$) & Gap & -0.0224 & 0.0129 & 0.0097 & 0.0406 & 0.0011 & 0.0807 \\
    \bottomrule
    \end{tabular}
\end{table*}

\subsubsection{Smoothing-stage choices and retained design}

The final stage examines strictly causal smoothing. This is the decisive step in both regimes. Table~\ref{tab::smoothing_tv_combined} reports changes in total variation relative to the no-smoothing baseline and shows a strong common pattern: all smoothing procedures reduce instability, but the magnitude of the reduction differs sharply across methods. In both the global and local settings, the largest gains are delivered by the Kalman-type smoothers, with the weighted arctanh Kalman specification producing the strongest or near-strongest reductions throughout the distribution of bank-level outcomes.

Those gains come with a responsiveness cost. Table~\ref{tab::smoothing_lag_combined} shows that the more aggressive Kalman-based methods shift the lag at which maximum alignment is attained, while Table~\ref{tab::smoothing_corr_combined} shows that they also attenuate maximum cross-correlation magnitude relative to the no-smoothing baseline. The same trade-off therefore appears in both regimes: stronger smoothing suppresses noise more effectively, but it also delays and attenuates the reaction to new information. Lighter EMA-based procedures preserve responsiveness better, yet they also deliver markedly smaller stability gains.

This is the central empirical result of the design study. Aggregation-stage choices and filling rules matter, but mostly at the margin. The dominant stabilizing mechanism in both pipelines is the final smoother. Because the reconstruction objective is to recover a stable latent sentiment process under sparse and irregular observation, stability is prioritized over immediate responsiveness, and the weighted arctanh Kalman filter is retained for both regimes. That retained choice should nevertheless be interpreted together with its lag cost when reading the downstream market-consistency evidence.

Table~\ref{tab:design_summary} summarizes the retained configurations. Two conclusions deserve emphasis. First, the preferred pipelines are not identical across regimes: the principal differences arise in the aggregation-stage weighting rules and in the filling specification. Second, the strongest cross-regime regularity is the dominance of causal smoothing. That commonality is more important than any isolated small difference among the earlier component choices and provides the clearest general lesson of the design study.

\begin{table*}[!ht]
\centering
\caption{Smoothing-stage comparisons, reported as differences in $\mathrm{TV}\!\left(S_S\right)$ relative to the no-smoothing baseline. Negative values indicate lower total variation than the baseline. Panel~A reports the global regime; Panel~B reports the local regime.}
\label{tab::smoothing_tv_combined}
\begin{tabular}{lrrrrrr}
\toprule
\multicolumn{7}{l}{\textbf{Panel A: Global aggregation}}\\
\midrule
Method & $\Delta$ median & $\Delta$ IQR & $\Delta$ mean & $\Delta$ std & $\Delta p_{10}$ & $\Delta p_{90}$ \\
\midrule
No smoothing (baseline) & 8.4382 & 10.2474 & 9.9976 & 7.6583 & 0.9759 & 20.6744 \\
EMA smoothing & -0.3063 & -0.3877 & -0.5573 & -0.5520 & -0.0982 & -1.2168 \\
Weighted EMA smoothing & -4.6362 & -5.5296 & -5.4652 & -4.2249 & -0.2917 & -12.2043 \\
Simple Kalman filter & -6.3433 & -7.9334 & -7.8394 & -6.2692 & -0.6597 & -16.6805 \\
Arctanh Kalman filter & -6.1701 & -7.4127 & -7.7142 & -6.0206 & -0.7004 & -16.4263 \\
Weighted Kalman filter & -7.0988 & -8.4640 & -8.5493 & -6.5953 & -0.7241 & -17.8910 \\
Beta-binomial smoother & -5.9923 & -8.8124 & -7.6715 & -6.4857 & -0.3972 & -16.9534 \\
Weighted arctanh Kalman filter & -7.6051 & -9.6136 & -9.1615 & -7.1027 & -0.7827 & -19.2596 \\
Adaptive-count Kalman & -5.0691 & -6.6171 & -6.3570 & -5.1967 & -0.1580 & -14.4604 \\
\midrule
\multicolumn{7}{l}{\textbf{Panel B: Local aggregation}}\\
\midrule
Method & $\Delta$ median & $\Delta$ IQR & $\Delta$ mean & $\Delta$ std & $\Delta p_{10}$ & $\Delta p_{90}$ \\
\midrule
No smoothing (baseline) & 7.8328 & 7.0352 & 8.3569 & 5.6778 & 2.7292 & 13.6330 \\
EMA smoothing & -2.1842 & -2.0970 & -2.1275 & -1.6132 & -0.6330 & -3.7399 \\
Weighted EMA smoothing & -2.9813 & -2.0365 & -2.8802 & -1.8984 & -1.1383 & -4.2483 \\
Simple Kalman filter & -6.9404 & -6.6558 & -7.3916 & -5.2632 & -2.1790 & -12.2861 \\
Arctanh Kalman filter & -6.9312 & -6.6653 & -7.3773 & -5.2584 & -2.1733 & -12.2675 \\
Weighted Kalman filter & -7.0076 & -6.6284 & -7.4680 & -5.2803 & -2.2267 & -12.3709 \\
Beta-binomial smoother & -6.8462 & -6.5393 & -7.3231 & -5.2228 & -2.1436 & -12.1866 \\
Weighted arctanh Kalman filter & -7.0719 & -6.6497 & -7.5286 & -5.3038 & -2.2197 & -12.4323 \\
Adaptive-count Kalman & -1.9341 & -1.6660 & -2.0597 & -1.4797 & -0.8480 & -2.7668 \\
\bottomrule
\end{tabular}
\end{table*}

\begin{table*}[!ht]
\centering
\caption{Lag at which maximum alignment is attained under alternative smoothing strategies. Panel~A reports the global regime; Panel~B reports the local regime.}
\label{tab::smoothing_lag_combined}
\begin{tabular}{lrrrrrr}
\toprule
\multicolumn{7}{l}{\textbf{Panel A: Global aggregation}}\\
\midrule
Method & median & IQR & mean & std & p10 & p90 \\
\midrule
No smoothing (baseline) & 0.0 & 0.0 & 0.0 & 0.0 & 0.0 & 0.0 \\
EMA smoothing & 0.0 & 0.0 & 0.0 & 0.0 & 0.0 & 0.0 \\
Weighted EMA smoothing & 0.0 & 0.0 & 0.1707 & 0.4363 & 0.0 & 1.0 \\
Simple Kalman filter & 2.0 & 1.0 & 1.6829 & 0.5604 & 1.0 & 2.0 \\
Arctanh Kalman filter & 2.0 & 1.0 & 1.6829 & 0.5604 & 1.0 & 2.0 \\
Weighted Kalman filter & 2.0 & 1.0 & 1.4634 & 0.7018 & 0.0 & 2.0 \\
Beta-binomial smoother & 0.0 & 0.0 & 0.1707 & 0.5366 & 0.0 & 0.0 \\
Weighted arctanh Kalman filter & 0.0 & 2.0 & 0.7805 & 0.9756 & 0.0 & 2.0 \\
Adaptive-count Kalman & 0.0 & 0.0 & 0.0976 & 0.3699 & 0.0 & 0.0 \\
\midrule
\multicolumn{7}{l}{\textbf{Panel B: Local aggregation}}\\
\midrule
Method & median & IQR & mean & std & p10 & p90 \\
\midrule
No smoothing (baseline) & 0.0 & 0.0 & 0.0488 & 0.3085 & 0.0 & 0.0 \\
EMA smoothing & 0.0 & 0.0 & 0.0 & 0.0 & 0.0 & 0.0 \\
Weighted EMA smoothing & 0.0 & 0.0 & 0.0 & 0.0 & 0.0 & 0.0 \\
Simple Kalman filter & 2.0 & 0.0 & 1.6585 & 0.2859 & 1.0 & 2.0 \\
Arctanh Kalman filter & 2.0 & 0.0 & 1.6585 & 0.2859 & 1.0 & 2.0 \\
Weighted Kalman filter & 2.0 & 1.0 & 1.5610 & 0.3496 & 1.0 & 2.0 \\
Beta-binomial smoother & 2.0 & 1.0 & 1.5366 & 0.3532 & 1.0 & 2.0 \\
Weighted arctanh Kalman filter & 2.0 & 0.0 & 1.6341 & 0.2939 & 1.0 & 2.0 \\
Adaptive-count Kalman & 0.0 & 0.0 & 0.0244 & 0.0329 & 0.0 & 0.0 \\
\bottomrule
\end{tabular}
\end{table*}

\begin{table*}[!ht]
\centering
\caption{Differences in maximum cross-correlation magnitude relative to the no-smoothing baseline. Negative values indicate lower peak-correlation magnitude than the baseline. Panel~A reports the global regime; Panel~B reports the local regime.}
\label{tab::smoothing_corr_combined}
\begin{tabular}{lrrrrrr}
\toprule
\multicolumn{7}{l}{\textbf{Panel A: Global aggregation}}\\
\midrule
Method & $\Delta$ median & $\Delta$ IQR & $\Delta$ mean & $\Delta$ std & $\Delta p_{10}$ & $\Delta p_{90}$ \\
\midrule
No smoothing (baseline) & 1.0 & 0.0 & 1.0 & 0.0 & 1.0 & 1.0 \\
EMA smoothing & -0.0103 & 0.0047 & -0.0098 & 0.0041 & -0.0151 & -0.0040 \\
Weighted EMA smoothing & -0.1595 & 0.0657 & -0.1566 & 0.0669 & -0.2277 & -0.0534 \\
Simple Kalman filter & -0.3077 & 0.1061 & -0.2971 & 0.1133 & -0.4207 & -0.0995 \\
Arctanh Kalman filter & -0.2981 & 0.1590 & -0.3050 & 0.1300 & -0.4582 & -0.1584 \\
Weighted Kalman filter & -0.4315 & 0.2358 & -0.4071 & 0.1763 & -0.5975 & -0.1743 \\
Beta-binomial smoother & -0.3808 & 0.2103 & -0.3902 & 0.2080 & -0.6164 & -0.1486 \\
Weighted arctanh Kalman filter & -0.5750 & 0.3027 & -0.5034 & 0.2396 & -0.7426 & -0.0966 \\
Adaptive-count Kalman & -0.2608 & 0.2113 & -0.2694 & 0.1784 & -0.5519 & -0.0075 \\
\midrule
\multicolumn{7}{l}{\textbf{Panel B: Local aggregation}}\\
\midrule
Method & $\Delta$ median & $\Delta$ IQR & $\Delta$ mean & $\Delta$ std & $\Delta p_{10}$ & $\Delta p_{90}$ \\
\midrule
No smoothing (baseline) & 1.0 & 0.0 & 1.0 & 0.0 & 1.0 & 1.0 \\
EMA smoothing & -0.0155 & 0.0044 & -0.0144 & 0.0043 & -0.0186 & -0.0100 \\
Weighted EMA smoothing & -0.0303 & 0.0156 & -0.0372 & 0.0259 & -0.0589 & -0.0223 \\
Simple Kalman filter & -0.3993 & 0.1171 & -0.3560 & 0.1347 & -0.4986 & -0.1231 \\
Arctanh Kalman filter & -0.3937 & 0.1199 & -0.3546 & 0.1337 & -0.5005 & -0.1281 \\
Weighted Kalman filter & -0.4091 & 0.1157 & -0.3841 & 0.1358 & -0.5322 & -0.1716 \\
Beta-binomial smoother & -0.3671 & 0.1300 & -0.3373 & 0.1271 & -0.4855 & -0.1376 \\
Weighted arctanh Kalman filter & -0.4234 & 0.0934 & -0.3934 & 0.1436 & -0.5363 & -0.1280 \\
Article-Count-Aware Kalman & -0.0190 & 0.0131 & -0.0261 & 0.0307 & -0.0512 & -0.0076 \\
\bottomrule
\end{tabular}
\end{table*}

\begin{table*}[!ht]
\centering
\caption{Retained pipeline configuration per aggregation regime, as identified by the sequential design study.}
\label{tab:design_summary}
\begin{tabularx}{\textwidth}{@{}Xll@{}}
\toprule
Component & Global & Local \\
\midrule
Uncertainty weighting & Polarity & None (baseline) \\
Redundancy control    & Corroboration (CC+Corr) & Deduplication (CC) \\
Time decay            & Exp.\ decay ($\tau{=}5$) & Exp.\ decay ($\lambda{=}0.95$) \\
Filling rule          & Constant fill & Const.-decay ($H{=}60$) \\
Smoother              & W.\ Kalman (arc) & W.\ Kalman (arc) \\
\bottomrule
\end{tabularx}
\end{table*}

\subsection{Market-consistency analysis}

To compare candidate sentiment-reconstruction pipelines, we evaluate each configuration under a common validation setup. Unless otherwise stated, the following parameter values are held fixed across all experiments:
\begin{itemize}
    \item maximum CCF lag: $4$ weeks,
    \item DTW rolling window: $52$ weeks,
    \item DTW step size: $4$ weeks,
    \item DTW Sakoe--Chiba band: $8$ weeks,
    \item DTW permutation count: $500$,
    \item Granger rolling window: $52$ weeks,
    \item Granger step size: $4$ weeks,
    \item Granger lag order: $2$,
    \item spectral segment length: $52$ weeks,
    \item Fisher aggregation weights: squared-size weights,
    \item Fisher aggregation method: weighted mean.
\end{itemize}

Each candidate configuration is evaluated twice: once under \emph{local} aggregation and once under \emph{global} aggregation, as defined in Section~\ref{sec:aggregation}. This ensures that any differences in consistency are attributable to the pipeline design itself rather than to a single aggregation scheme.

For every configuration, we compute the same family of consistency diagnostics introduced in Section~\ref{sec:experiment}: 
(i) the peak prewhitened cross-correlation between sentiment and price, 
(ii) the share of rolling windows in which sentiment Granger-causes price, 
(iii) the medium-run coherence fraction, and 
(iv) the mean rolling DTW-Pearson statistic, which is treated as a supplementary diagnostic rather than a primary validation metric.

To rank configurations, we define a composite consistency score as a weighted average of the normalized validation components:
\begin{equation}
\label{eq:consistency_score}
\mathrm{Score}
=
0.4\,\widetilde{\rho}_{\mathrm{CCF}}
+
0.3\,\widetilde{G}_{S\rightarrow Y}
+
0.2\,\widetilde{C}_{\mathrm{mid}}
+
0.1\,\widetilde{r}_{\mathrm{DTW}},
\end{equation}
where:
\begin{itemize}
    \item $\widetilde{\rho}_{\mathrm{CCF}}$ is the normalized peak prewhitened cross-correlation aggregate,
    \item $\widetilde{G}_{S\rightarrow Y}$ is the normalized proportion of rolling windows with significant sentiment-to-price Granger causality,
    \item $\widetilde{C}_{\mathrm{mid}}$ is the normalized medium-run coherence fraction,
    \item $\widetilde{r}_{\mathrm{DTW}}$ is the normalized mean rolling DTW-Pearson statistic.
\end{itemize}

All quantities are z-score normalized before aggregation so that no single metric dominates the composite score purely because of scale. Higher values of $\mathrm{Score}$ indicate that the reconstructed sentiment signal exhibits stronger and more stable consistency with market data across the primary validation dimensions.

The weighting scheme in Eq.~\eqref{eq:consistency_score} assigns the greatest importance to the prewhitened CCF, since it provides the cleanest direct measure of lead--lag association after removing within-series autocorrelation. Rolling Granger causality is weighted second, as it captures directional incremental predictability. Spectral coherence receives a lower but still material weight because it measures whether the sentiment--price relationship is concentrated in the medium-run band of interest. Finally, DTW receives the smallest weight because, as discussed in Section~\ref{sec:dtw}, it is used only as a supplementary robustness diagnostic and does not carry the same inferential status as the primary methods.

The best-performing candidate is therefore the configuration that maximizes Eq.~\eqref{eq:consistency_score} while maintaining methodological plausibility across both local and global aggregation settings.

\begin{table*}[!ht]
\centering
\begin{tabular}{lrrrrrrr}
\toprule
& Banks & CCF lag & CCF $\bar{\rho}$ & Granger S$\to$P (\%) & Mid-coh. & DTW $\bar{r}$ & Score \\
\midrule
$S_{159}$ & 37 & 3 & 0.0223 & 8.0952 & 0.3501 & 0.5731 & 0.9743 \\
$S_{151}$ & 37 & 3 & 0.0223 & 8.0952 & 0.3501 & 0.5731 & 0.9743 \\
$S_{447}$ & 37 & 3 & 0.0250 & 7.1429 & 0.3484 & 0.5729 & 0.8586 \\
$S_{439}$ & 37 & 3 & 0.0250 & 7.1429 & 0.3484 & 0.5729 & 0.8586 \\
$S_{291}$ & 37 & 3 & 0.0245 & 10.0317 & 0.3383 & 0.5322 & 0.7979 \\
$S_{435}$ & 37 & 3 & 0.0287 & 8.5873 & 0.3409 & 0.5328 & 0.7672 \\
$S_{258}$ & 37 & 3 & 0.0377 & 8.8348 & 0.3351 & 0.5293 & 0.7329 \\
$S_{373}$ & 37 & 3 & 0.0440 & 8.1325 & 0.3360 & 0.5087 & 0.7208 \\
$S_{293}$ & 37 & 3 & 0.0321 & 8.1325 & 0.3414 & 0.4955 & 0.7007 \\
$S_{181}$ & 37 & 3 & 0.0407 & 8.1325 & 0.3365 & 0.5093 & 0.6946 \\
\midrule
$S_{165}$ & 37 & 3 & 0.0292 & 8.5847 & 0.3519 & 0.5122 & 0.6128 \\
$S_{414}$ & 37 & 3 & 0.0423 & 7.4074 & 0.3484 & 0.4919 & 0.5609 \\
$S_{398}$ & 37 & 3 & 0.0366 & 10.0529 & 0.3369 & 0.4828 & 0.5438 \\
$S_{446}$ & 37 & 3 & 0.0316 & 11.0053 & 0.3340 & 0.4874 & 0.5327 \\
$S_{366}$ & 37 & 3 & 0.0422 & 7.4074 & 0.3475 & 0.4861 & 0.5295 \\
$S_{206}$ & 37 & 3 & 0.0337 & 10.5291 & 0.3358 & 0.4771 & 0.5245 \\
$S_{270}$ & 37 & 3 & 0.0415 & 8.1481 & 0.3446 & 0.4669 & 0.5150 \\
$S_{174}$ & 37 & 3 & 0.0425 & 7.0370 & 0.3478 & 0.4807 & 0.4846 \\
$S_{357}$ & 37 & 3 & 0.0276 & 7.7513 & 0.3509 & 0.5149 & 0.4731 \\
$S_{222}$ & 37 & 3 & 0.0399 & 7.4074 & 0.3464 & 0.4838 & 0.4675 \\
\bottomrule
\end{tabular}
\caption{Top-ranked sentiment-reconstruction pipelines under global aggregation, evaluated using the market-consistency diagnostics introduced in Section~\ref{sec:experiment}. The upper block reports results with respect to rolling weekly price differences, while the lower block reports results with respect to the 90-day rolling min--max scaled price series. Higher composite scores indicate stronger joint consistency across prewhitened cross-correlation, rolling Granger causality, medium-run coherence, and DTW alignment.}
\label{tab::signal_study_global}
\end{table*}

\begin{table*}[!ht]
    \centering
    \begin{tabular}{lrrrrrrr}
\toprule
& Banks & CCF lag & CCF $\bar{\rho}$ & Granger S$\to$P (\%) & Mid-coh. & DTW $\bar{r}$ & Score \\
\midrule
$S_{192}$ & 37 & 3 & 0.0325 & 8.1491 & 0.3184 & 0.6853 & 1.0284 \\
$S_{32}$ & 37 & 3 & 0.0322 & 8.1491 & 0.3186 & 0.6831 & 1.0228 \\
$S_{208}$ & 37 & 3 & 0.0394 & 7.6186 & 0.3183 & 0.6868 & 1.0177 \\
$S_{400}$ & 37 & 3 & 0.0388 & 7.6186 & 0.3180 & 0.6867 & 1.0025 \\
$S_{384}$ & 37 & 3 & 0.0317 & 8.1491 & 0.3176 & 0.6858 & 1.0003 \\
$S_{352}$ & 37 & 3 & 0.0360 & 7.6186 & 0.3187 & 0.6860 & 0.9822 \\
$S_{240}$ & 37 & 3 & 0.0303 & 7.7181 & 0.3189 & 0.6829 & 0.9257 \\
$S_{304}$ & 37 & 3 & 0.0323 & 5.8945 & 0.3173 & 0.6849 & 0.5837 \\
$S_{448}$ & 37 & 3 & 0.0356 & 5.4635 & 0.3179 & 0.6858 & 0.5622 \\
$S_{256}$ & 37 & 3 & 0.0343 & 5.4635 & 0.3180 & 0.6853 & 0.5466 \\

\midrule
$S_{314}$ & 37 & 3 & 0.0338 & 7.5545 & 0.3252 & 0.6641 & 1.0740 \\
$S_{367}$ & 37 & 3 & 0.0293 & 7.8065 & 0.3296 & 0.6727 & 1.0248 \\
$S_{366}$ & 37 & 3 & 0.0340 & 7.2031 & 0.3242 & 0.6627 & 0.9959 \\
$S_{170}$ & 37 & 3 & 0.0318 & 7.3630 & 0.3262 & 0.6669 & 0.9767 \\
$S_{362}$ & 37 & 3 & 0.0311 & 7.5545 & 0.3264 & 0.6676 & 0.9674 \\
$S_{175}$ & 37 & 3 & 0.0285 & 7.8065 & 0.3292 & 0.6717 & 0.9581 \\
$S_{266}$ & 37 & 3 & 0.0283 & 8.6013 & 0.3281 & 0.6649 & 0.9562 \\
$S_{271}$ & 37 & 3 & 0.0281 & 7.7874 & 0.3296 & 0.6722 & 0.9391 \\
$S_{174}$ & 37 & 3 & 0.0332 & 7.2031 & 0.3238 & 0.6621 & 0.9332 \\
$S_{410}$ & 37 & 3 & 0.0247 & 8.6013 & 0.3331 & 0.6672 & 0.8938 \\
\bottomrule
\end{tabular}
\caption{Top-ranked sentiment-reconstruction pipelines under local aggregation, evaluated using the market-consistency diagnostics introduced in Section~\ref{sec:experiment}. The upper block reports results with respect to rolling weekly price differences, while the lower block reports results with respect to the 90-day rolling min--max scaled price series. Higher composite scores indicate stronger joint consistency across prewhitened cross-correlation, rolling Granger causality, medium-run coherence, and DTW alignment.}
\label{tab::signal_study_local}
\end{table*}

Tables~\ref{tab::signal_study_global} and \ref{tab::signal_study_local} report composite consistency scores for all evaluated configurations under global and local aggregation respectively. The full pipeline compositions are listed in the Appendix (Tables~\ref{tab::strat_names_global} and \ref{tab::strat_names_local}).

\paragraph{Headline finding: the three-week lag.}

\textbf{Across every reported configuration, the peak prewhitened cross-correlation is attained at a lag of exactly three weeks}, even though lags up to four weeks were considered. This holds irrespective of aggregation regime, price transformation, and pipeline design. This structural regularity---a consistently preferred alignment horizon rather than a distribution of lag values across configurations---is arguably the
most informative result of the consistency study. It suggests that the detected sentiment--price relationship is not incidental to any particular specification but reflects a stable temporal structure in the data.

\paragraph{Magnitude and ranking.}
The absolute relationship is modest: average CCF values are uniformly low, and the proportion of rolling windows with significant Granger causality is limited. The composite score functions less as evidence of strong predictability and more as a \emph{filtering device}: it eliminates configurations that produce unstable or incoherent signals and retains those that reproduce the same weak but consistent sentiment--price structure across multiple validation dimensions.

Local aggregation yields a somewhat higher cluster of top scores and a tighter spread; global aggregation remains competitive but more dispersed. This difference should not be interpreted as local aggregation uncovering a stronger sentiment effect; rather, it suggests that separating thematic streams before aggregation preserves the limited available signal more reliably.

\subsection{Overall summary}

The four components of the results section cohere into a single narrative:
\begin{enumerate}
  \item The raw data exhibits exactly the structural pathologies (sparsity, irregularity, redundancy, heteroscedastic coverage) that motivate the reconstruction framework.
  \item The impulse test confirms strict causality across all configurations; the duplicate test shows that redundancy control materially improves robustness.
  \item The sequential design study identifies component choices that are empirically justified, not arbitrary: the preferred configuration (Table~\ref{tab:design_summary}) follows from the data's structural features.
  \item The market-consistency analysis reveals a three-week lead--lag pattern, confirming that carefully reconstructed signals exhibit structured plausibility relative to observable market dynamics, even when the absolute predictive signal is weak.
\end{enumerate}

\section{Limitations}
\label{sec:limitations}

Several limitations should be acknowledged when interpreting these results.

\paragraph{Title-only data.}  The dataset consists of article titles rather
than full content. Titles capture the primary framing of a story but may
omit contextual details that influence sentiment. Extending the pipeline to
full-text representations is a natural direction for future work.

\paragraph{Fixed classifier.}  The sentiment classifier is treated as a
fixed black-box component. Errors or systematic biases in its probability
vectors propagate into the reconstructed signal in ways the pipeline cannot
correct~\cite{battaglia2024inference}; the framework provides no mechanism to estimate or bound this
propagation. Jointly learning the classifier and reconstruction procedure
remains an open problem.

\paragraph{Domain and generalizability.}  The empirical analysis focuses
on AI-related news for a selected set of financial firms. While this
provides a controlled case study, the results may not generalize to other
sectors, languages, or news sources without further validation. The
label-free evaluation framework, however, is domain-agnostic.

\paragraph{Weak price signal.}  The observed sentiment--price relationship
is weak by design: AI-related news is one of many factors influencing large
financial institutions. The modest magnitude of consistency metrics is
expected and does not reflect a limitation of the reconstruction framework
per se.

\paragraph{Small panel and approximate inference.}  The analysis panel
comprises 43 Banks. Per-bank estimates are noisy, cross-bank aggregates
inherit that uncertainty, and Fisher-combined $p$-values are approximate
due to plausible cross-bank dependence under shared macroeconomic exposures.

\section{Conclusion}
\label{sec:conclusion}

This paper argues that sentiment estimation from sparse news data is better
framed as a \emph{causal signal reconstruction} problem than as a
classification challenge. We propose a modular three-stage pipeline---
aggregation with uncertainty and redundancy awareness, causal gap-filling,
and causal smoothing---and introduce a label-free evaluation framework for
settings where ground-truth longitudinal labels are unavailable.

The key empirical finding is a robust three-week lead--lag structure between
reconstructed AI-news sentiment and stock prices, stable across all tested
pipeline configurations and aggregation regimes. The sequential design
study further reveals that causal smoothing is the single most effective
reconstruction component, that the preferred configuration differs modestly
between global and local aggregation, and that redundancy control has
measurable practical value beyond its theoretical motivation.

Together, these results demonstrate that careful signal reconstruction
substantially improves the reliability of text-based sentiment indicators.
The contribution is not a new classifier but a principled engineering
framework for turning classifier outputs into deployable temporal signals.

Future work may extend the framework in several directions: incorporating
full-text rather than title-level representations; modeling event-level
sentiment dynamics rather than title-level aggregates; jointly estimating
the classifier and reconstruction procedure to propagate calibration
uncertainty; applying the label-free evaluation methodology to non-financial
sparse NLP settings such as regulatory monitoring or clinical trial sentiment.

\bibliographystyle{ieeetr}
\bibliography{bibliography}

\appendix
\section{Pipeline Composition Tables}
\label{sec:appendix_tables}

Tables~\ref{tab::strat_names_global} and \ref{tab::strat_names_local} list
the full composition of each evaluated pipeline configuration.

\noindent\textbf{Abbreviation legend:}
\begin{itemize}[noitemsep]
  \item \textbf{CC}: connected-component duplicate detection with
        downweighting ($\phi_{\mathrm{ded}}$, full normalisation).
  \item \textbf{CC+Corr}: connected-component detection with
        corroboration boost ($\phi_{\mathrm{cor}}$).
  \item \textbf{W.\ Kalman (arc)}: weighted arctanh Kalman filter.
\end{itemize}

\begin{table*}[!hb]
\centering
    \begin{tabular}{llllll}
\toprule
 & Aggregator & Fill rule & Smoother & Weighting \\
\midrule
$S_{159}$ & Weighted Exp. Decay ($\tau= 5$) & Const.-decay fill & Beta-binomial & -- \\
$S_{151}$ & Weighted Exp. Decay ($\tau= 5$) & Const. proj. & Beta-binomial & -- \\
$S_{447}$ & Weighted Exp. Decay ($\tau= 5$) & Const.-decay fill & Beta-binomial & Polarity, CC+Corr. \\
$S_{439}$ & Weighted Exp. Decay ($\tau= 5$) & Const. proj. & Beta-binomial & Polarity, CC+Corr. \\
$S_{291}$ & Weighted Exp. Decay ($\tau= 5$) & Const. proj. & Kalman (arctanh) & Polarity, CC \\
$S_{435}$ & Weighted Exp. Decay ($\tau= 5$) & Const. proj. & Kalman (arctanh) & Polarity, CC+Corr. \\
$S_{258}$ & Weighted Exp. Decay ($\lambda = 0.95$)  & Const. proj. & Kalman & Polarity, CC \\
$S_{373}$ & Weighted Exp. Decay ($\alpha= 0.95$) & Const. proj. & Weighted Kalman & Entropy, CC+Corr. \\
$S_{293}$ & Weighted Exp. Decay ($\tau= 5$) & Const. proj. & Weighted Kalman & Polarity, CC \\
$S_{181}$ & Weighted Exp. Decay ($\alpha= 0.95$) & Const. proj. & Weighted Kalman & Entropy \\
\midrule
$S_{165}$ & Weighted Exp. Decay ($\lambda = 0.95$)  & Const. proj. & Weighted Kalman & Entropy \\
$S_{414}$ & Weighted Exp. Decay ($\lambda = 0.95$)  & Const.-decay fill & Weighted Kalman (arc) & Polarity, CC+Corr. \\
$S_{398}$ & Weighted Exp. Decay ($\tau= 5$) & Const.-decay fill & Weighted Kalman (arc) & Entropy, CC+Corr. \\
$S_{446}$ & Weighted Exp. Decay ($\tau= 5$) & Const.-decay fill & Weighted Kalman (arc) & Polarity, CC+Corr. \\
$S_{366}$ & Weighted Exp. Decay ($\lambda = 0.95$)  & Const.-decay fill & Weighted Kalman (arc) & Entropy, CC+Corr. \\
$S_{206}$ & Weighted Exp. Decay ($\tau= 5$) & Const.-decay fill & Weighted Kalman (arc) & Entropy \\
$S_{270}$ & Weighted Exp. Decay ($\lambda = 0.95$)  & Const.-decay fill & Weighted Kalman (arc) & Polarity, CC \\
$S_{174}$ & Weighted Exp. Decay ($\lambda = 0.95$)  & Const.-decay fill & Weighted Kalman (arc) & Entropy \\
$S_{357}$ & Weighted Exp. Decay ($\lambda = 0.95$)  & Const. proj. & Weighted Kalman & Entropy, CC+Corr. \\
$S_{222}$ & Weighted Exp. Decay ($\lambda = 0.95$)  & Const.-decay fill & Weighted Kalman (arc) & Polarity \\
\bottomrule
\end{tabular}
\caption{Pipeline composition of top-ranked configurations under global aggregation. Upper block: evaluated against rolling weekly log returns. Lower block: evaluated against 90-day rolling log returns.}
\label{tab::strat_names_global}
\end{table*}

\begin{table*}[!hb]
    \centering
    \begin{tabular}{llllll}
\toprule
 & Aggregator & Fill rule & Smoother & Weighting \\
\midrule
$S_{192}$ & Weighted Exp. Decay ($\alpha= 0.95$) & Const.-decay fill & Weighted EMA & Entropy \\
$S_{32}$ & Weighted Mean & Const.-decay fill & Weighted EMA & Entropy \\
$S_{208}$ & Weighted Exp. Decay ($\tau= 5$) & Const.-decay fill & Weighted EMA & Entropy \\
$S_{400}$ & Weighted Exp. Decay ($\tau= 5$) & Const.-decay fill & Weighted EMA & Entropy, CC+Corr. \\
$S_{384}$ & Weighted Exp. Decay ($\alpha= 0.95$) & Const.-decay fill & Weighted EMA & Entropy, CC+Corr. \\
$S_{352}$ & Weighted Exp. Decay ($\tau= 5$) & Const.-decay fill & Weighted EMA & Entropy, CC \\
$S_{240}$ & Weighted Exp. Decay ($\alpha= 0.95$) & Const.-decay fill & Weighted EMA & Polarity \\
$S_{304}$ & Weighted Exp. Decay ($\tau= 5$) & Const.-decay fill & Weighted EMA & Polarity, CC \\
$S_{448}$ & Weighted Exp. Decay ($\tau= 5$) & Const.-decay fill & Weighted EMA & Polarity, CC+Corr. \\
$S_{256}$ & Weighted Exp. Decay ($\tau= 5$) & Const.-decay fill & Weighted EMA & Polarity \\
\midrule
$S_{314}$ & Weighted Exp. Decay ($\lambda = 0.95$) & Const.-decay fill & Kalman & Entropy, CC \\
$S_{367}$ & Weighted Exp. Decay ($\lambda = 0.95$) & Const.-decay fill & Beta-binomial & Entropy, CC+Corr. \\
$S_{366}$ & Weighted Exp. Decay ($\lambda = 0.95$) & Const.-decay fill & W. Kalman (arc) & Entropy, CC+Corr. \\
$S_{170}$ & Weighted Exp. Decay ($\lambda = 0.95$) & Const.-decay fill & Kalman & Entropy \\
$S_{362}$ & Weighted Exp. Decay ($\lambda = 0.95$) & Const.-decay fill & Kalman & Entropy, CC+Corr. \\
$S_{175}$ & Weighted Exp. Decay ($\lambda = 0.95$) & Const.-decay fill & Beta-binomial & Entropy \\
$S_{266}$ & Weighted Exp. Decay ($\lambda = 0.95$) & Const.-decay fill & Kalman & Polarity, CC \\
$S_{271}$ & Weighted Exp. Decay ($\lambda = 0.95$) & Const.-decay fill & Beta-binomial & Polarity, CC \\
$S_{174}$ & Weighted Exp. Decay ($\lambda = 0.95$) & Const.-decay fill & W. Kalman (arc) & Entropy \\
$S_{410}$ & Weighted Exp. Decay ($\lambda = 0.95$) & Const.-decay fill & Kalman & Polarity, CC+Corr.\\
\bottomrule
\end{tabular}
\caption{Pipeline composition of top-ranked configurations under local aggregation. Upper block: evaluated against rolling weekly log returns. Lower block: evaluated against 90-day rolling log returns.}
\label{tab::strat_names_local}
\end{table*}

\end{document}